\documentclass[journal]{IEEEtran}
\ifCLASSINFOpdf
\else
\fi
\usepackage[cmex10]{amsmath}
\usepackage[pdftex]{graphicx}
\usepackage{graphicx}
\usepackage{float}
\usepackage{array}
\usepackage{amsmath}
\usepackage{cases}
\usepackage{verbatim}
\usepackage{lipsum}
\usepackage{siunitx}
\usepackage[]{cite}
\usepackage[caption=false,font=footnotesize]{subfig}
\usepackage{wrapfig}
\usepackage{dblfloatfix}
\usepackage[symbol]{footmisc}
\usepackage{multirow}

\begin{document}
\title{The Utility of Synthetic Reflexes and Haptic Feedback for Upper-Limb Prostheses\\in a Dexterous Task Without Direct Vision}

\author{Neha Thomas$^{1}$, Farimah Fazlollahi$^{2}$, Katherine J. Kuchenbecker$^{2}$, and Jeremy D. Brown$^{3}$% <-this % stops a space
\thanks{$^{1}$Neha Thomas is with the Department of Biomedical Engineering, Johns Hopkins School of Medicine, and the Haptic Intelligence Department, Max Planck Institute for Intelligent Systems, {\tt\small neha.thomas@jhmi.edu}}% <-this % stops a space
 
\thanks{$^{2}$Farimah Fazlollahi and Katherine J. Kuchenbecker are with the Haptic Intelligence Department, Max Planck Institute for Intelligent Systems, {\tt\small fazlollahi@is.mpg.de} and {\tt\small kjk@is.mpg.de}}
        
\thanks{$^{3}$Jeremy D. Brown is with the Department of Mechanical Engineering, Johns Hopkins University, {\tt\small jdelainebrown@jhu.edu}}%
}

\maketitle

\begin{abstract}
Individuals who use myoelectric upper-limb prostheses often rely heavily on vision to complete their daily activities. They thus struggle in situations where vision is overloaded, such as multitasking, or unavailable, such as poor lighting conditions. Non-amputees can easily accomplish such tasks due to tactile reflexes and haptic sensation guiding their upper-limb motor coordination. Based on these principles, we developed and tested two novel prosthesis systems that incorporate autonomous controllers and provide the user with touch-location feedback through either vibration or distributed pressure. These capabilities were made possible by installing a custom contact-location sensor on the fingers of a commercial prosthetic hand, along with a custom pressure sensor on the thumb. We compared the performance of the two systems against a standard myoelectric prosthesis and a myoelectric prosthesis with only autonomous controllers in a difficult reach-to-pick-and-place task conducted without direct vision. Results from 40 non-amputee participants in this between-subjects study indicated that vibrotactile feedback combined with synthetic reflexes proved significantly more advantageous than the standard prosthesis in several of the task milestones. In addition, vibrotactile feedback and synthetic reflexes improved grasp placement compared to only synthetic reflexes or pressure feedback combined with synthetic reflexes. These results indicate that both autonomous controllers and haptic feedback facilitate success in dexterous tasks without vision, and that the type of haptic display matters. 
\end{abstract}


% \fi
\IEEEpeerreviewmaketitle

\begin{figure}[t]
		\centering
		%\vspace{1em}
% 		\includegraphics[width=\columnwidth]{Figures/Hand_V2.pdf}
		\includegraphics[width=\columnwidth]{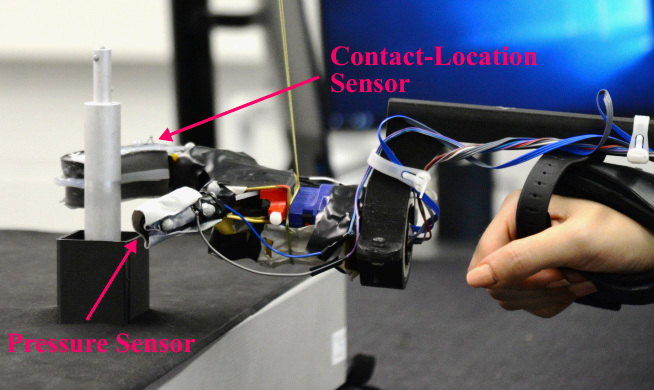}
% 		\includegraphics[width=\columnwidth]{Figures/Hand_V5.pdf}
%		\includegraphics[width=\columnwidth]{Figures/Hand_V6.pdf}
	   % \vspace{-1.75em}
		\caption{Non-amputee participants wore a myoelectric prosthesis via a wrist-brace adaptor to pick and place a cylindrical object without directly observing the interaction. Custom tactile sensors fitted to the thumb and fingers were used to implement reflexive controller strategies and provide haptic feedback.}
		\label{fig:hand}
% 		\vspace{-1em}
\end{figure}

\section{Introduction}
\IEEEPARstart{I}{n} dexterous reach-to-lift-and-grasp tasks without vision, non-amputee individuals use tactile sensations from their hand and fingers to localize and form their hand to the contours of the object upon contact in order to securely yet economically grasp it \cite{Karl2013NonvisualReach,Winges2003TheGrasp,Gentilucci1997TactileMovements}. Haptic (tactile, kinesthetic, and proprioceptive) cues are then used to inform motor coordination to successfully lift the object \cite{Johansson1996SensoryHumans}. In addition, reflexive control is induced in response to cutaneous signals indicating tactile events such as slippage or unanticipated deformation of the object. These reactive sensorimotor controllers compensate for grip-force errors by facilitating rapid grip-force adjustments \cite{Johansson1992Sensory-MotorActions,Nakajima2006Location-specificMuscles} and thus serve to complement volitional motor control. 

Amputees using clinical myoelectric prostheses lack the haptic sensation that is essential for dexterous sensorimotor control. Instead, they must rely heavily on vision to complete activities of daily living \cite{Atkins1996EpidemiologicPriorities,Sobuh2014}. This visual dependency is not only cognitively burdensome \cite{Thomas2020}, but it also significantly limits manipulation abilities in activities where vision is constrained or unavailable (e.g., watching a screen, searching for an object in the dark). 
Thus, in order for an amputee to dexterously accomplish a reach-to-grasp-and-lift task with their prosthesis in the absence of vision, the prosthesis must support volitional and reflexive control in a manner consonant with the intact sensorimotor system. In particular, it should include support for: (1) tactile sensing capable of detecting contact location; (2) haptic feedback mechanisms for conveying contact information to the amputee; (3) an ability to reflexively react to adverse events, like slip or excessive grasping force, which could unintentionally deform or break objects.

Various approaches for sensing contact location for robotic hands and fingers have been discussed in the literature. One common approach is to create individual tactile sensing elements arranged in an array or matrix \cite{Ponraj2019ActiveGrippers,Osborn2014}. While this taxel-based approach is capable of measuring pressure and contact location, it requires many sensing elements to cover a large area, and the sensed location is discrete. Furthermore, increasing the resolution of the system requires reducing the size of the taxels as well as the distance between them, which can be difficult to construct. Electronic skins that do not require as many sensing elements as a typical tactile array can provide continuous, multi-site contact location but generally require substantial computational power to compute accurate measures \cite{Lee2021Piezo}. Finally, there are some commercial sensors like the BioTac \cite{Jimenez2014} that have been used for sensing in prosthetics, but they provide tactile data only for the fingertip, which may be insufficient for ensuring a stable whole-hand grasp, and they are typically expensive and delicate. 
% In contrast, we previously introduced a novel contact-location sensor that uses only three electronic leads and provides continuous, single-site contact location over a relatively large area \cite{Thomas2021Sensorimotor-inspiredVision}. 

Considerable research has also focused on the challenge of providing haptic feedback of contact location \cite{Antfolk2013b,Hartmann2014TowardsEmbodiment,Shehata2020MechanotactileUse}. Often, these approaches have been limited to discrete location feedback and involve the use of multiple mechanotactile actuators. Researchers have, for example, used both servo motors and vibrotactile actuators mounted on the forearm to portray contact location from each of the five fingers of a prosthesis \cite{Antfolk2013b, Antfolk2013a}. These methods, however, cannot provide continuous feedback of contact location and also tend to be bulky.  Electrotactile feedback has also been investigated to provide discretized contact location and force \cite{Hartmann2014TowardsEmbodiment,Ward2018Multi-channelHand, Scott1980Sensory-feedbackControl}. While it may be more compact than mechanical actuators, the stimulation from electrotactile feedback has been shown to interfere with EMG signals and elicit sensations that can be perceived as unpleasant \cite{Antfolk2013SensoryProsthetics}.
% In our previous work, we provided continuously amplitude- and pattern-modulated vibrotactile feedback to convey continuous contact location and two discrete pressures. 

Paralleling advancements in haptic sensing and haptic feedback technologies for upper-limb prostheses are investigations into the efficacy and utility of autonomous control approaches for prosthetic hands \cite{Salisbury1967APrehension,Chappell1987ControlHand,Nightingale1985MicroprocessorArm}. Researchers have shown that slip prevention and compliant grasping controllers reduce object slips and breaks 
\cite{Osborn2016, Matulevich2013UtilityControl}. Similar controllers have even been implemented in commercial hands such as the Ottobock SensorHand Speed \cite{OttobockSensorHand}. In addition to improving functional performance, these controllers alleviate some of amputees' mental burden, as they operate without the user in the control loop. However, these controllers can still fail because of false negatives or false positives. In the former scenario, the controller misses an adverse event, while in the latter, an unwanted reaction is generated (e.g., increasing grip force during a fragile object transfer). Both these failure modes could cause the user to distrust the system because they are unaware of the contexts or reasons for failure. 

In an effort to overcome these limitations on contact-location sensing, contact-location feedback, and autonomous control, we previously developed a sensorimotor-inspired prosthesis system featuring a novel contact-location sensor and vibrotactile feedback with anti-slip and anti-overgrasping reflex controllers \cite{Thomas2021Sensorimotor-inspiredVision}. Our contact-location sensor uses only three electrical leads and provides continuous, single-site contact location over the outer and inner surfaces of the fingers. Additionally, we provided continuously amplitude- and frequency-modulated vibrotactile feedback to convey continuous contact location and the presence of an object in the grasp, as sensed by the thumb-mounted pressure sensor. We showed that the combination of vibrotactile feedback and reflexive control in a myoelectric prosthesis improved performance consistency of a reach-to-pick-and-place task without direct vision \cite{Thomas2021Sensorimotor-inspiredVision} compared to performance with a standard myoelectric prosthesis. 

While our prior work demonstrated the potential utility of a sensorimotor-inspired prosthesis control system, there are still many knowledge gaps that need to be addressed to advance such a system towards clinical viability. First, the sensitivity and resolution of the contact-location sensor need to be thoroughly characterized to inform future research. Second, it should be determined whether the addition of haptic feedback offers a significant improvement over pure reflexive control. Third, a modality-matched haptic feedback approach should be investigated to see whether it offers a significant improvement over non-modality-matched vibrotactile display. Indeed, modality-matched feedback is thought to be more intuitive and feel more natural to the average user \cite{Kim2010OnProsthetics}. Thus, we advance our prior work here with: (1) a characterization of the contact-location sensor, (2) an additional experimental condition consisting of the myoelectric prosthesis with only tactile reflexes, and (3) an additional experimental condition of the myoelectric prosthesis with tactile reflexes and modality-matched distributed pressure feedback of contact location. We hypothesize that the particular combination of reflex controllers and distributed pressure feedback would result in the largest improvement over the standard prosthesis, due to the modality-matched contact-location feedback. With this work, we aim to provide additional contexts for how a hybrid approach may work to improve prosthesis performance. 
% Tactile sensors literature

% Haptic feedback of contact location literature

% Reflex controllers literature

% Previous study on haptic feedback

% Haptic feedback has been used to display force, pressure, temperature...etc. few tsudies have invest
% discretized haptic feedback contact location

% reflex controllers

% 1) tasks done without vision, multitasking

% 2)prosthetic limbs, reliance on vision
%     cognitively demanding
%     undesriable at times
%     unavailable

% 3)substituting haptics for vision
%     cognitively better
%     performance similar
%     variety of types of haptic feedback
% 4) reflex controllers
%     less burden
%     improves performance
% 5) demand for reflex     controllers and haptic feedback
% for situations low vision
% 6) previous study
\begin{figure}[t]
		\centering
		\vspace{1em}
		\includegraphics[width=\columnwidth]{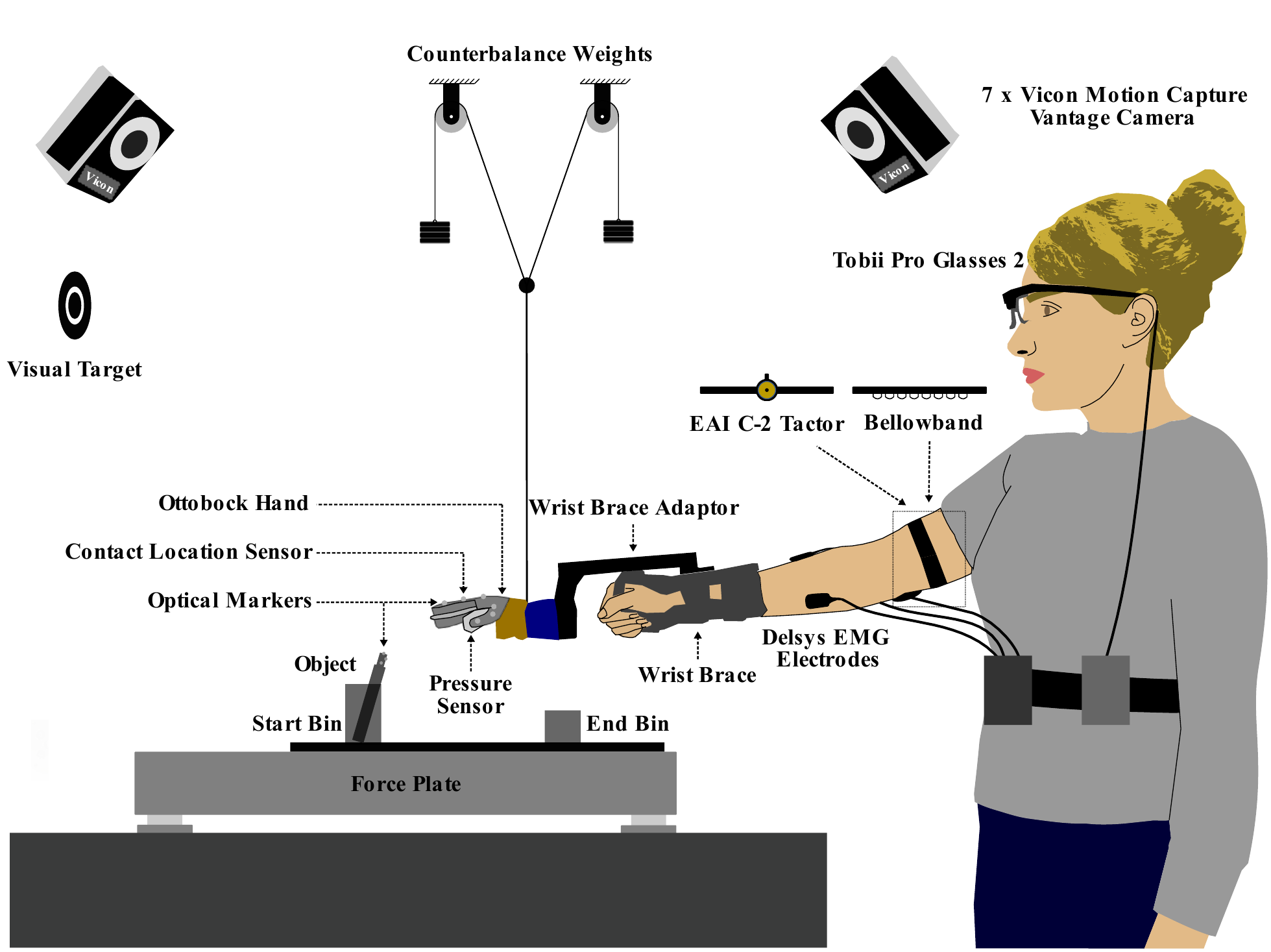}
	   % \vspace{-1.75em}
		\caption{The experiment involves picking up and moving a cylindrical object using a myoelectric prosthesis fitted with two custom sensors. Furthermore, participants had to fixate on a visual target in front of them, rather than on the interaction of the prosthetic hand and the object. Peripheral vision was not occluded. In addition to receiving aid from reflex controllers in the prosthesis, some participants also received haptic feedback in the form of either vibrotactile feedback (C-2 tactor) or distributed pressure (Bellowband).}
		\label{fig:setup}
% 		\vspace{-1em}
\end{figure}
\section{Methods}

\subsection{Participants}
Under the MPI-IS Haptic Intelligence Department's framework agreement with the Max Planck Ethics Council (protocol number F005D, approved in February of 2021), we recruited 31 new participants 
% 48 participants (13 female, 35 male, age 31.4 $\pm$ 6.68) 
to perform a reach-to-pick-and-place task using a myoelectric prosthesis in a between-subjects study with four experimental conditions. Data from 17 participants from our previous study \cite{Thomas2021Sensorimotor-inspiredVision} were re-used in the analysis for the present study; 8 were in the Standard condition, and 9 were in the condition with both vibrotactile feedback and tactile reflexes. In total, there were 48 participants (13 female, 35 male, age 31.4 $\pm$ 6.68).
% All 47 participants ( provided informed consent under the Haptic Intelligence Department’s framework agreement protocol number F005D. 
Participants were randomly assigned to one of four conditions that were balanced for gender and handedness; the (self-reported) five left-handed participants and one ambidextrous participant all did the study with the right hand. The experiment lasted approximately one hour, and participants not employed by the Max Planck Society received 8 euros per hour as compensation. 

\subsection{Experimental Task}
Participants used the myoelectric prosthesis to grasp and relocate a cylindrical aluminum object (12 cm long, 2 cm diameter) from one fixed bin ($3.8 \times 3.8 \times 7.6$ cm) to another stationary bin ($3.8 \times 3.8 \times 5.1$ cm) that was 17.5 cm away (see \ref{fig:setup}. This object roughly resembles the size and shape of a thick highlighter pen or an electric toothbrush. Additionally, participants were required to complete said task without looking directly at the prosthetic hand or object. Rather, they had to fixate on a visual target 3\,m away on the wall in front of them (peripheral vision was not occluded). The eye-tracking glasses provided a measure of the participants' exact gaze direction. This visual constraint mimics multitasking situations where visual attention is diverted away from a dexterous task, such as when grabbing a cup of tea while focusing on a video presentation. 
% Using the myoelectric prosthesis, participants transported a cylindrical object (12 cm long, 2 cm diameter) situated initially inside a small, stationary bin ($3.8 \times 3.8 \times 7.6$ cm) to a second stationary bin ($3.8 \times 3.8 \times 5.1$ cm) 17.5 cm away. The dimensions of this object are similar to common household items like a marker or a screwdriver. During this reach-to-pick-and-place task, participants were also required to avoid looking at the object. Instead, participants looked at a target 3\,m away on the wall in front of them. This visual constraint was implemented to mimic a multi-tasking situation where visual attention is diverted away from a dexterous task, such as when grabbing a cup of tea while attending a Zoom meeting.
The absence of direct vision rendered this reach-to-pick-and-place task especially difficult in two important ways. First, due to the slim profile of the cylindrical object and the geometry of the hand, the object should be grasped at an ideal location with the correct orientation. Second, excessive grasping force caused the object to slide out of hand's grasp. Thus, it was hypothesized that haptic feedback of contact location and the presence of the object in the grasp, in combination with autonomous grasping controllers, would assist in these two challenges.

\subsection{Experimental Hardware and Software}

The measurement devices in the system included a seven-camera Vicon Vantage motion-capture system, a custom-built three-axis force plate, and Tobii Pro 2 eye-tracking glasses to identify the participant's gaze direction. A Delsys Bagnoli surface electromyography (sEMG) system was used for proportional myoelectric control of the Ottobock SensorHand Speed prosthesis using two sEMG electrodes on the wrist flexor and extensor muscle groups. Two custom-built tactile sensors were placed on the thumb and finger of the prosthesis (see Fig. \ref{fig:hand}). Control was implemented through an NI myRIO DAQ and Simulink with QUARC Real-Time software at a 1000\,Hz sampling rate. Complete details of our measurement hardware are presented in \cite{Thomas2021Sensorimotor-inspiredVision}.

The two haptic feedback displays were a C-2 tactor to provide vibrotactile feedback (driven by NI myRIO and a linear current amplifier) and an eight-tactor Bellowband pneumatic display \cite{Young2019Bellowband:Vibration} to provide distributed pressure feedback. The Bellowband was programmed in C and controlled with an NI cDAQ-9174 housing an analog input module (NI-9205) and an analog output module (NI-9264) at a 250\,Hz sample rate.

The 1-DoF Ottobock SensorHand Speed myoelectric prosthesis was worn by able-bodied individuals using a 3D-printed adaptor attached to a wrist brace. A counterweight pulley system was implemented to offset 80\% of the prosthesis's mass (500\,g) to replicate the load typically experienced by amputees. 
The entire setup is shown in Fig. \ref{fig:setup}.
% The opening and closing of the hand was driven proportionally using signals from the wrist extensor and flexor muscle groups, respectively. Complete details of the EMG control approach and calibration procedures can be found in \cite{Thomas2021Sensorimotor-inspiredVision}.
% The experimental setup is depicted in Fig. \ref{fig:setup}.

\subsection{Tactile Sensors}
Two custom-built fabric-based sensors were used to separately obtain pressure and contact-location information. The piezoresistive pressure sensor was similar to the one developed by Osborn et al. \cite{Osborn2016} and was placed on the prosthesis thumb. It operates on the same principles as a force-sensitive resistor sensor. Our novel contact-location sensor \cite{Thomas2021Sensorimotor-inspiredVision} was wrapped around the fingers, covering both palmar and dorsal regions. The contact-location sensor consists of two layers, both of which are fixed separately within a silicone frame. The bottom layer consists of a long piece of piezoresistive fabric, while the top layer consists of a long piece of conductive fabric. When a voltage is applied across the length of the piezoresistive fabric, a voltage gradient is created; when the top conductive layer contacts the bottom at a specific point, a distinct output voltage is generated, similar to a potentiometer. For a depiction of the sensor layers, refer to Fig. 3 in \cite{Thomas2021Sensorimotor-inspiredVision}.
% (see Fig. \ref{fig:sensor}). 
% The contact-location sensor detects single-site points of contact and is similar in function to a moving voltage divider - pressing this sensor at a specific point elicits a certain voltage. 
% can we reuse the figures from IROS?

\begin{figure}[t]
		\centering
		%\vspace{1em}
		\includegraphics[width=\columnwidth]{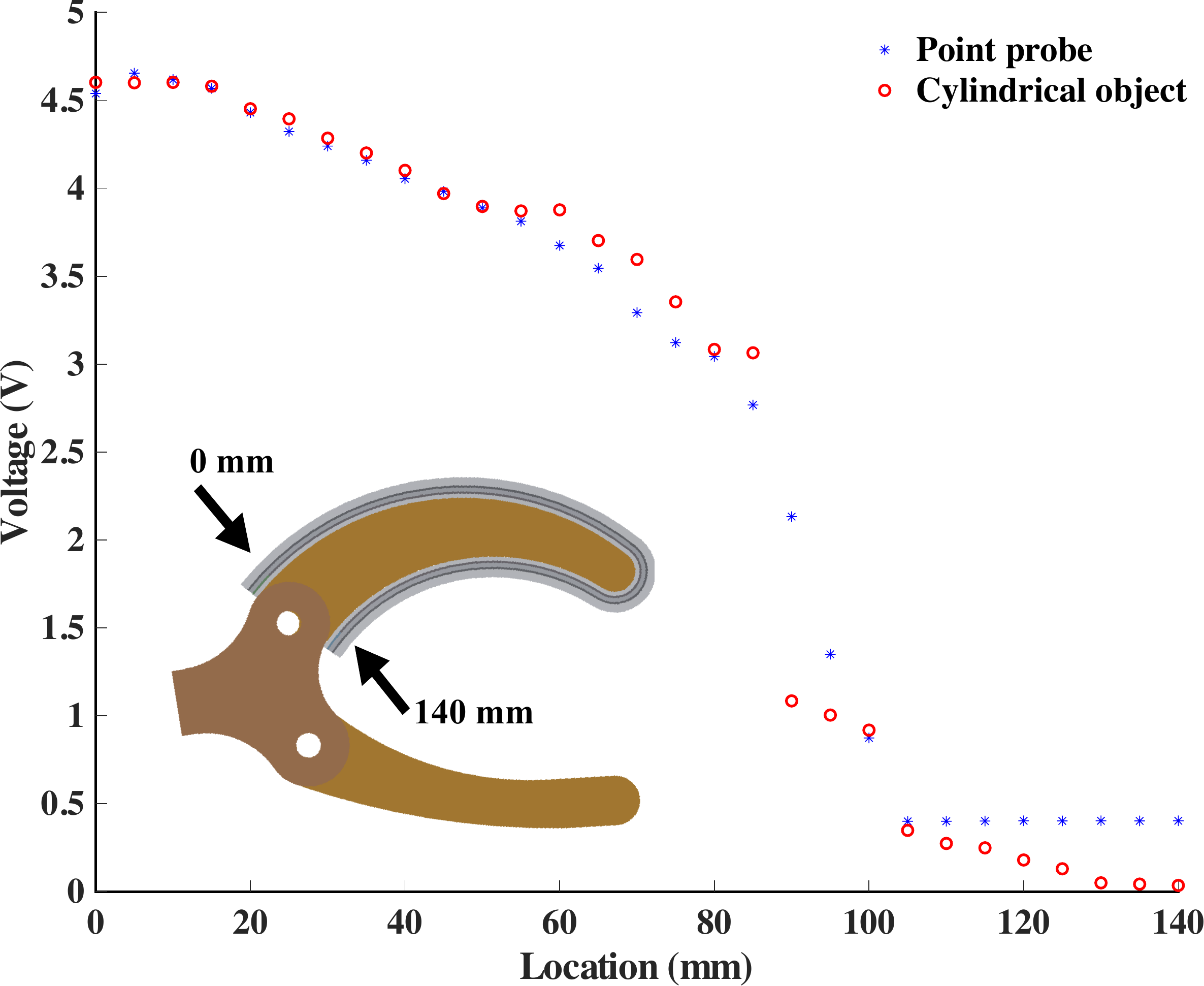}
	   % \vspace{-1.75em}
		\caption{The voltage output of the contact-location sensor when pressed with a point probe and a cylindrical probe the size of the test object. 0\,mm is the most proximal dorsal location, while 140\,mm is the most proximal palmar location.} %(b) the minimum force required to activate the sensor using a flat probe and a cylindrical probe at various locations (represented here by the voltage elicited). The dashed line indicates the mean, while the solid lines indicate the standard deviation.}
		\label{fig:characterization}
% 		\vspace{-1em}
\end{figure}

The relationship between contact location along the length of the sensor and output voltage is shown in Fig. \ref{fig:characterization}a: the sensor's response was characterized for both a 2.6\,mm by 0.6\,mm point probe and a cylindrical object whose dimensions matched the test object in our experiments. The cylindrical object was oriented perpendicular to the sensor, as though it was being grasped. The mapping exhibits nonlinear behavior due to the concavity in the inner region of the prosthesis fingers. When the contact-location sensor was not pressed, the baseline voltage reading was around 0.4\,V. This is the lower limit for contact with the point probe, which occurs around 105\,mm. In contrast, the cylindrical probe is able to elicit lower voltages because it makes contact with a larger area of the sensor. To determine the minimum force required to activate the sensor, we used an ATI Nano 17 force/torque sensor with both the cylindrical probe and a 17\,mm circular flat probe to press at 29 evenly distributed locations. The average activation force when using the cylindrical probe was 1.5 $\pm$ 0.55 N, and it was 2.2 $\pm$ 0.56 N for the flat probe. 
%(see Fig. \ref{fig:characterization}b.

% add the characeterization figures
\subsection{Haptic Feedback Systems}
The haptic feedback is designed to provide two kinds of information that are important for grasping accuracy: 1) where contact occurs on the prosthesis fingers, and 2) whether an object is in the grasp of the prosthesis. 
\subsubsection{Vibrotactile Feedback}
\indent The C-2 tactor was worn just above the elbow on the biceps muscle. As in \cite{Thomas2021Sensorimotor-inspiredVision}, it provided amplitude-modulated feedback of contact location and frequency-modulated binary feedback of grasping pressure. 

\begin{figure}[t]
		\centering
		\vspace{1em}
		\includegraphics[width=\columnwidth]{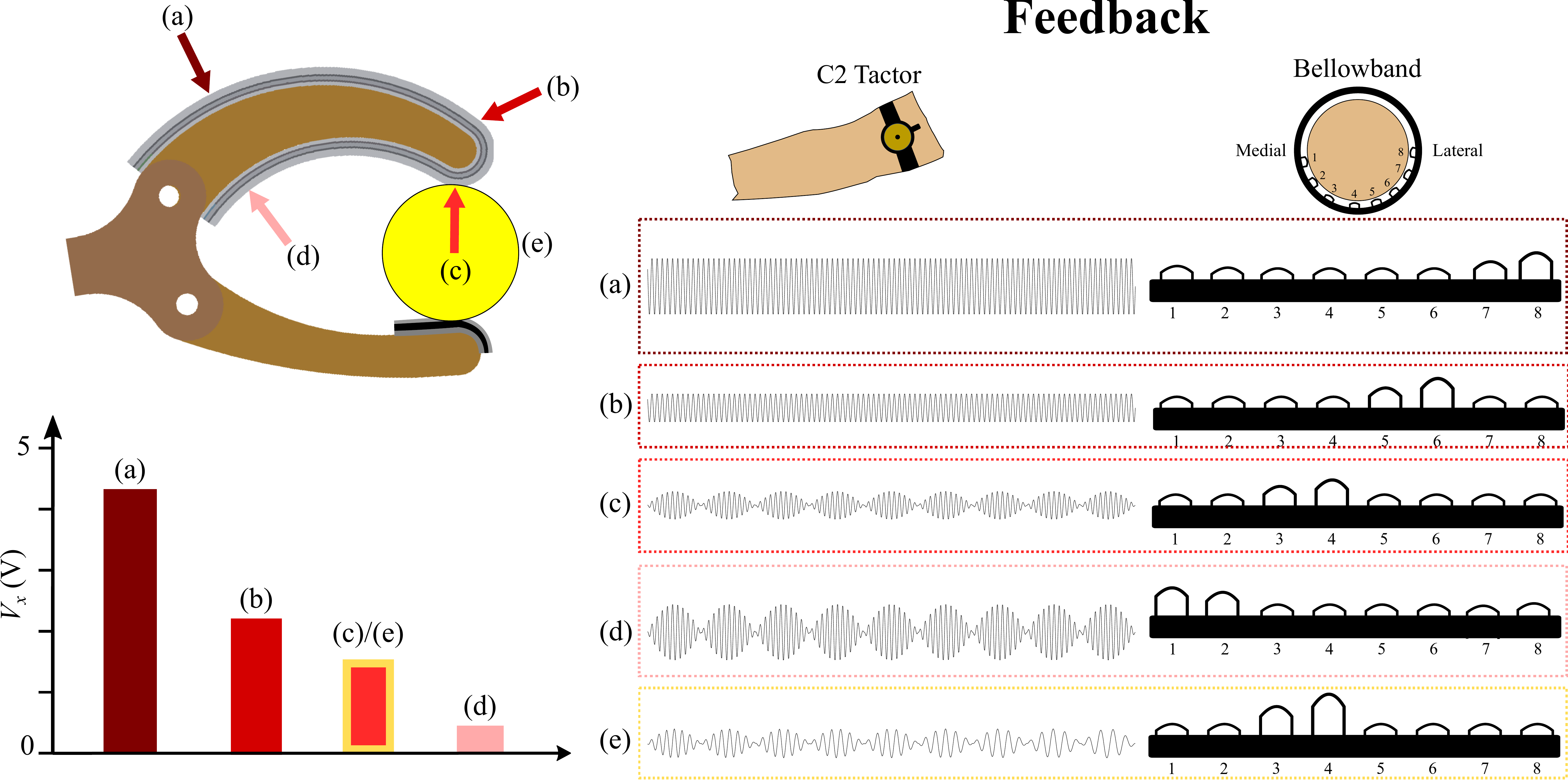}
	   % \vspace{-1.75em}
		\caption{
% 		From the top left clockwise -- contact-location sensor: a) and b) show the silicone frames of the top and bottom layers of the sensor, respectively. Item c) is the long layer of conductive fabric; d) and e) are the conductive fabric electrodes attached to the two ends of f) the lower piezoresistive fabric layer, which has a voltage gradient across its length. The sensor is wrapped around the curved external surface of the fingers of g) the prosthesis. When something touches the outside of the sensor, the top layer of conductive fabric touches the bottom layer of piezoresistive fabric at the contact point. 
Left: Example voltages measured at different points on the contact-location sensor. (a) A proximal dorsal location. (b) A distal dorsal location. (c) A distal palmar location. (d) A proximal palmar location. (e) An object grasped by the prosthesis; it touches the same distal palmar point contacted in (c). Right: Feedback signals for the vibrotactile and pneumatic pressure actuators corresponding to the four example contact locations (a--d); the depicted vibration signals are 1\,s long, and (e) starts 1\,s after object contact. As shown in (e), the haptic feedback is altered distinctively when both the pressure sensor and the contact-location sensor are activated, since this condition indicates an object is most likely within the grasp of the prosthesis.}
		\label{fig:sensor}
% 		\vspace{-1em}
\end{figure}

The contact-location sensor's signals were first normalized between 0 (proximal) and 1 (distal), regardless of whether contact was on the dorsal or palmar region. Contact on the dorsal side was mapped to a constant vibration, while contact on the palmar side was mapped to a pulsed vibration. The mapping equation for current input to the C-2 tactor was
\begin{equation}
I = \begin{cases}
     A(x) \cdot \sin{(W \cdot 250\,\textrm{Hz} \cdot t)} &, \,  \text{dorsal}\\ 
    E_v(t) \cdot f(x) \cdot \sin{(W \cdot f \cdot t)} &, \,  \text{palmar}\\
  \end{cases}
  \label{Eqn:VibMapping}
\end{equation}
\noindent where $x$ is the normalized contact-location sensor signal, $A(x)$ is $0.5\, \textrm{A} \cdot \sqrt{1-x}$, $W$ is $2 \pi\,\frac{\textrm{rad}}{\textrm{cycle}}$, $E_v(t)$ is an envelope function denoted by $|\sin{(W \cdot 4.75\,\textrm{Hz} \cdot t)}|$, $f$ is the frequency in Hz, and $t$ is the time in seconds. Example signals are shown in Fig. \ref{fig:sensor}. When the pressure sensor on the prosthesis thumb exceeds a heuristically determined threshold ($p_g>0.2$\,V signifying object contact), the frequency of the vibration stimulus decreases linearly from 250\,Hz to 150\,Hz over a 2\,s period, as shown in \ref{fig:sensor}(e).\\

\subsubsection{Pneumatic Pressure Feedback}
% \indent Pneumatic feedback of contact location was provided by Bellowband, a band consisting of 8 pneumatic actuators (bellows), which was first presented in \cite{Young2019Bellowband:Vibration}. 
Each bellow of the Bellowband represented a different region of the contact-location sensor. The Bellowband was worn on the upper arm, with the bellow representing the fingertip located at the posterior part of the arm, above the elbow. This orientation of the Bellowband was chosen because the two-point discrimination threshold is smaller on the posterior part of the upper arm compared to the anterior part \cite{Nolan1982Two-PointWomen,Koo2016}, which we confirmed in pilot studies. At most, two neighboring bellows were activated to indicate transitions between regions. 

%Algorithm equations. maybe it is more clear if an example is shown (edges vs middle)
\begin{figure}[t]
		\centering
		\vspace{1em}
		\includegraphics[width=\columnwidth]{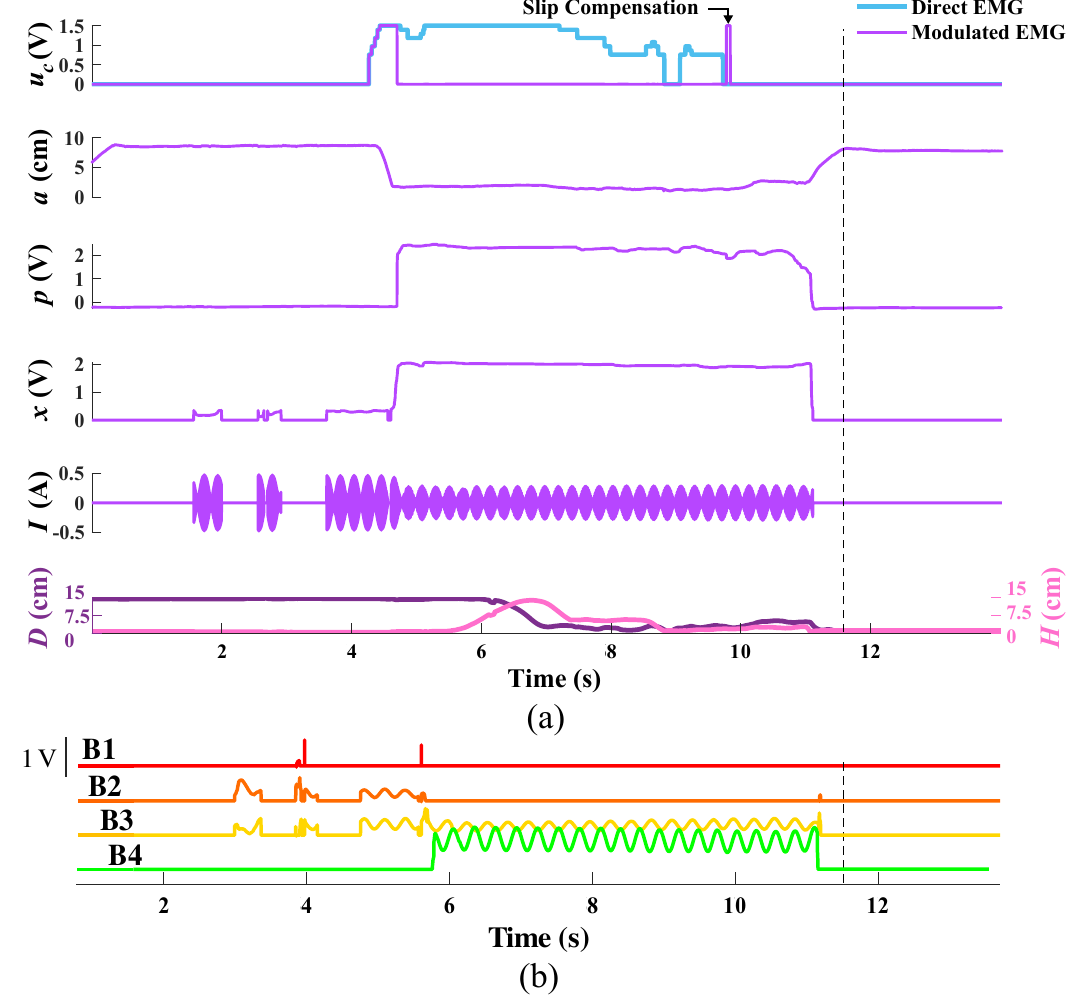}
	   % \vspace{-1.75em}
		\caption{(a) Excerpt of time-series traces from a representative participant's trial in the Reflex-Vib condition as they found, grasped, picked up, moved, and set down the object. The vertical dashed line indicates the time point when the participant successfully placed the object into the end bin. The traces shown are the closing command $u_c$, the grip aperture $a$, the pressure sensor signal $p$, the contact-location sensor signal $V_x$, the C-2 tactor signal $I$, and the object's displacement $D$ from the end bin and height $H$ above the force plate as measured by the motion-capture system. The participant first attempts to localize the prosthesis hand on the object, as shown by the contact-location signal and C-2 current traces. Next, the participant activates their EMG, which is modulated once the pressure sensor signal ramps up. One fast slip event is also detected from the pressure sensor signal and compensated for. (b) Activation signals for four of the bellows on the pneumatically actuated pressure band as they would be driven by the pressure sensor and contact-location sensor signals from (a). Bellows 5, 6, 7, and 8 (not shown) would be completely deflated at a constant 0.1\,V because all contacts occurred on the interior surface of the fingers in this trial.}
		\label{fig:Traces}
		\vspace{-1em}
\end{figure}
To map the detected contact location to the bellows on the Bellowband, $B_1$, $B_2$, $\cdots$, $B_8$, the sensor output voltage was delineated into eight regions. Defining voltages $\nu_1$, $\nu_2$, $\cdots$, $\nu_9$ were used to demarcate the boundaries between sensor regions, such that any voltage $V_x$ elicited by the sensor falls between the two consecutive defining voltages $\nu_i$ and $\nu_{i+1}$ ($i$ $\in$ 1...8), which 
corresponds to bellow $B_i$. We command the pressure profile $P_i(t)$ to this bellow as follows:
\begin{equation}
    P_i(t) = E_p(t) \cdot \left [(p_{\max}-p_{\min}) \cdot \gamma_i  + p_{\min} \right]
    \label{Eqn: pressure bband}
\end{equation}
\noindent where $E_p(t)={0.25\cdot\sin{(2\pi\frac{\textrm{rad}}{\textrm{cycle}} \cdot 3\,\textrm{Hz} \cdot t)}+0.75}$ is an envelope function that was used to prevent sensory adaptation to the pressure stimuli, $p_{\max}$ is the maximum allowable pressure, $\gamma_i$ is a proportional gain (defined in Eqn. \ref{eqn_gamma}) that represents where $V_x$ falls between $\nu_i$ and $\nu_{i+1}$, and $p_{\min}=0.1$\,V is the minimum pressure (completely deflated bellow). We set $p_{\max}=0.8$\,V (partially inflated) when providing only contact location feedback from the contact location sensor. Alternatively, when the pressure sensor on the prosthesis thumb exceeds a heuristically determined threshold ($p_g>0.2$\,V signifying object grasp), we set $p_{\max}=1.5$\,V (completely inflated). This larger maximum pressure value differentiates the simultaneous activation of the pressure sensor and contact-location sensor from just the contact-location sensor alone, similar to the change of the vibration stimulus frequency. 

Pilot testing showed that discrete jumps between neighboring bellows were difficult to interpret, so we developed a method for distributing actuation between two neighboring bellows. For each sensor region, we defined a threshold voltage $\tau_{i}$, where $\nu_i < \tau_{i} < \nu_{i+1}$. If $\nu_i < V_x < \tau_{i}$, neighboring bellow $B_{i-1}$ was activated in addition to $B_i$. Otherwise, neighboring bellow $B_{i+1}$ was activated. The proportion that each bellow is actuated depends on the gain $\gamma_i$, which is calculated as
\begin{equation}
  \gamma_i =  \hspace{-.3em}
  \begin{cases}
    0.5 + 0.5 \frac{V_x - \tau_{i}}{\nu_{i+1}-\tau_i}, & \hspace{-.5em} 
          \begin{cases} 
                i=1,\cdots,7 ; V_x \geq \tau_{i}\\

          \end{cases} \\
    0.5 + 0.5 \frac{V_x - \nu_{i}}{\tau_i - \nu_i}, &
    \hspace{-.5em}
         \begin{cases}
              i=2,\cdots,8; V_x < \tau_{i}\\
        
         \end{cases}\\
    1, & \hspace{-.5em}
        \begin{cases}
              i=1; V_x < \tau_{i} \\
              \text{or } i=8; V_x \geq \tau_{i} \\
        \end{cases}
  
  \end{cases}
%   \hspace{-1.5em}
  %\right\}
  \hspace{-2em}
  \label{eqn_gamma}
\end{equation}   
\noindent The pressure $P_{i\pm1}$ of the closer neighboring bellow $B_{i\pm1}$ is set to % neighboring bellow $B_{i\pm1}$ was activated according to the following laws:
\begin{equation}
    P_{i\pm1} = E_p(t) \cdot \left [(p_{\max}-p_{\min}) \cdot \gamma_{i\pm1} + p_{\min} \right]
\end{equation}
\noindent  where $\gamma_{i+1}$ is equal to $1-\gamma_i$ for $i = 1,...,7 \text{ \& } V_x \geq \tau_{i}$, and $\gamma_{i-1}$ is equal to $1-\gamma_i$ for $i=2,...,8 \text{ \& } V_x < \tau_{i}$.
% \begin{equation}
%     pr_{i+1} = 1-pr_i, \  i = 1,...,7 \text{ \& } V_x > q^x_{i,i+1}
%       \label{Eqn:ProportionBbandN+1}  
%   \end{equation}
%  
%
%   \begin{equation}
%     pr_{i-1} = 1-pr_i, \  i = 2,...,8 \text{ \& } V_x < q^x_{i,i+1}
%       \label{Eqn:ProportionBbandN+1}  
%     %   \label{Eqn:ProportionBbandN-1}  
%   \end{equation}
%  
% Similarly to (\ref{Eqn: pressure bband}), the pressure of neighboring bellows $P_{i+1}(t)$ or $P_{i-1}(t)$ were denoted by $E_2(t) \cdot \left [(p_{max}-p_{min}) \cdot pr_{i+1} + p_{min} \right]$ and $E_2(t) \cdot \left [(p_{max}-p_{min}) \cdot pr_{i-1} + p_{min} \right]$, respectively.
%
Fig. \ref{fig:sensor} shows the location of each bellow relative to the upper arm, as well as example actuation outputs.

% insert figure depicting bellow actuation

\subsection{Reflex System}
The reflex system consisted of three autonomous controllers to comprehensively prevent various grasp errors such as over-grasping of the objects, high-speed slips, and slow-speed slips. These controllers build on work done by Osborn  et al. \cite{Osborn2016} and rely on the pressure signal from the piezoresistive pressure sensor on the prosthetic thumb.

\subsubsection{Over-grasp Controller}
\indent This controller uses the pressure sensor signal to prevent excessive grasp force by modulating the closing command $u_{c}$ to the motor according to the control law
\begin{equation}
u_{c} = \begin{cases}
      u_{c} \cdot e^{-K \cdot p} &, \, p \geq p_{g}, \  \ \text {palmar}\\ 
    u_{c} &, \, \text {otherwise}\\
  \end{cases}
  \label{Eqn:AntiOverGrasp}
\end{equation}
\noindent where $K$ is $3$ V$^{-1}$, $p$ is the pressure sensor voltage, and $p_{g} = 0.2$\,V is the pressure threshold for detecting object contact.

\subsubsection{Anti-slip Controller}
This controller uses the pressure sensor signal to detect and respond to fast and slow slips. 

{\em Fast slips} were detected by rapid decreases in pressure according to the following equation:
\begin{equation}
{\text{Slip}}_{f} = \begin{cases}
      1  &, \, \frac{dp}{dt} \le {q}_{fs}\\
     0 &, \, \text {otherwise}\\
   \end{cases}
  \label{Eqn:Fastlip}
\end{equation}
\noindent where $\frac{dp}{dt}$ is the time derivative of the pressure sensor signal and $q_{fs}=-20$ Vs$^{-1}$ is a heuristically determined threshold. %a negative threshold.
When a fast slip occurs, a closing command is sent to the motor at maximum voltage for 60\,ms to prevent the object from falling out of the hand. 

{\em Slow slips} were detected by moderate decreases in pressure according to the following equation:
\begin{equation}
{\text{Slip}}_{s} = \begin{cases}
      1  &, \, p(t) - p(t-0.5\,\textrm{s}) < {p}_{ss} \\ 
      0 &, \, \text {otherwise}\\
  \end{cases}
  \label{Eqn:SlowSlip}
\end{equation}
\noindent where $p(t)$ is the pressure sensor signal at the current time and $p_{ss}= -0.35$\,V is a heuristically determined threshold.
When a slow slip occurs, a closing command is sent to the motor at maximum voltage for 30\,ms. See Fig. \ref{fig:Traces} for excerpts of relevant signals including the haptic feedback and slip control.

\subsection{Experimental Protocol}

% \vspace{0.5em}
% \subsubsection{Experimental Procedure}
% \indent  
Participants were randomized into one of four conditions: Standard (myoelectric prosthesis with no additional features), Reflex (prosthesis featuring reflex controllers), Reflex-Vib (prosthesis featuring reflex controllers and vibrotactile feedback), and Reflex-Pneu (prosthesis featuring reflex controllers and pneumatic pressure feedback). 
% Tactile sensors were on the prosthetic fingers and thumb for all four conditions, but were not active for the Standard condition. 
Each participant completed the task in only one of the four conditions (between-participants design). 

Because the eye-tracking glasses are not compatible with prescription glasses, all participants were required to successfully read the largest (topmost) line of an eye exam chart from a distance of 3~m before proceeding with the experiment.
Next, participants completed a demographics survey with questions regarding occupation, age, gender, handedness, and experience with myoelectric and haptic devices.

The experimenter then helped the participant don the prosthesis via the wrist-brace attachment. The participant's skin was cleaned with an alcohol wipe in preparation for the sEMG electrode placement. 
% One electrode was placed over the wrist flexor muscle group, and the other was placed over the wrist extensor muscle group. 
sEMG signals were calibrated using maximum voluntary contractions of the wrist flexor and extensor. For more details on calibration steps, refer to \cite{Thomas2021Sensorimotor-inspiredVision}.

Following calibration, the participant practiced controlling the prosthetic hand using their muscle activity.
% ; the magnitude of wrist flexion was proportional to closing speed of the hand, while the magnitude of wrist extension was proportional to the opening speed (see equation ?? in \cite{Thomas2021Sensorimotor-inspiredVision} for complete details). 
To account for typical EMG drift \cite{Kyranou2018CausesProstheses} that could occur during the experiment, participants were instructed on how to re-zero their signals. Participants re-zeroed their signals whenever they wished and also when prompted by the experimenter. 

If the participant was assigned to a condition receiving haptic feedback (Reflex-Vib or Reflex-Pneu), the proper device (C-2 tactor or the Bellowband) was attached to their upper arm. Finally, the experimenter helped the participant don the eye-tracking glasses. Calibration of the glasses was done through iMotions software. 

The experimenter then trained the participant on the ideal reach-to-pick-and-place strategy. After this coaching, participants were asked to complete the task successfully two times while being able to observe the prosthetic hand and object. They were then given 5 minutes to try to complete the task while looking only at the visual target on the wall. This timed practice session ended early if they successfully completed the task twice. Participants then completed twenty trials of the reach-to-pick-and-place task while keeping their gaze on the visual target. A trial began when the cylindrical object was placed inside the start bin, and it ended when the object was placed into the end bin or when 60 seconds had passed. 
After all twenty trials, participants completed a survey based on the NASA-TLX. Survey questions are described in Section \ref{Sec:survey metrics}.

\subsection{Metrics}

\subsubsection{Task Success}
To evaluate success in the reach-to-pick-and-place task, the following three milestones with binary outcomes were extracted from each trial: (1) successfully lifting the object from the start bin, (2) successfully reaching the end bin with the object, and (3) successfully setting the object inside the end bin. A lift is defined as holding the object in the air for at least 1 second. Reaching the end bin is defined as coming within an 8\,cm radius of the end bin. Motion-capture and early trial-completion data were used to track milestone achievement. The time required to reach each of the milestones was also measured. 
We also counted the number of drops that occurred after the object was successfully lifted. This number was determined by assessing sharp decreases in the object's height (relative to the prosthesis) using motion-capture data.

% \subsubsection{Interaction forces}
% For each trial, the integral of the sum of squares of the lateral forces detected by the force plate was calculated. Similarly, the integral of the absolute value of the normal force was measured for each trial. These values were normalized by the trial time. 

\subsubsection{Grasping Location}
To obtain the most reliable ground truth measurements of grasping location, the angle of the prosthesis fingertip relative to the object during attempted grasping was calculated using motion-capture data. These angles were measured for 1.75\,s before the completion of grasping. 

\subsubsection{Proportion of Time Spent Looking at the Task (Cheating)}
Gaze direction was analyzed using the eye-tracking data recorded in iMotions software. A horizontal line below the visual target was drawn for each frame of an individual's point-of-view recording. The gaze direction was automatically computed by iMotions. The proportion of time spent looking toward the task was calculated by dividing the time spent fixating below the horizontal line by the trial time. 

\subsubsection{Survey}
\label{Sec:survey metrics}
The post-experiment survey asked participants to rate their perceived performance at: (1) finding the object, (2) grasping the object, (3) lifting the object, (4) moving the object to the end bin, and (5) setting the object inside the end bin. It further asked them to rate their perceived mental effort, physical effort, and level of physical comfort during the experiment. Next it asked them to evaluate how much they relied on auditory, visual, and somatosensory cues to complete the task. Each of the rating questions was a sliding scale from 0 to 100. Finally, the survey prompted participants to provide comments and suggestions about their experience.

\subsection{Statistical Analysis}
All statistical tests were performed in RStudio v1.2.1335. For all mixed-model analysis, participant was treated as a random effect.  We use $\alpha=0.05$ to determine significance. We report the estimates of the fixed effects $\beta$ and their standard error $SE$. 

\subsubsection{Time Spent Fixating on the Task (Cheating)}
A linear mixed model was used to gauge differences in the proportion of time spent visually cheating among the conditions. The fixed effect was condition.

\subsubsection{Task Milestones}
Three separate logistic mixed-effects models were used to analyze the binary outcomes of lifting the object, reaching the end bin with the object, and setting down the object into the end bin. The fixed effects for these models were condition, the proportion of time spent looking toward the task (cheating), and trial number. These models were run on trials where participants looked away no more than 37\% of the time. This threshold was chosen as a compromise value halfway between the 75th percentile (24\%) and 50\% to balance removing too many and too few trials. This threshold penalized only those trials for which participants looked away from the visual target more than 37\% of the time, and the resulting dataset contained 701 out of 800 possible trials. 20 of the removed trials were from a participant for whom the eye-tracking system failed to record, while the remaining 79 trials removed were those who cheated more than 37\% of the time. 

Three separate linear mixed models were run to assess the time required to lift the object, move the object to the end bin, and set the object into the end bin. The fixed and random effects were the same as the previously mentioned logistic mixed models.

\subsubsection{Number of Drops}
A linear mixed model was used to assess the effect of condition and trial on the number of drops that occurred per trial. 
% In addition to 37\% frequency of looking away threshold, 
This model was run only on the 647 trials in which the object was lifted, regardless of cheating. %557 data points. 
% Similar linear mixed models were analyzed for the normalized lateral and normal forces.

\subsubsection{Grasp Location}
Finally, all successful grasps were analyzed to determine what the successful range of grasping locations was. A logistic mixed model was used on all grasps to assess the fixed effect of condition on whether the grasp location was within the successful range or not. In addition, the earth mover's distance metric was calculated to understand how the histograms of successful grasping locations compared to the histograms of all grasping locations for each condition.

\subsubsection{Survey}
Separate linear models were run for each of the rating questions described in Section \ref{Sec:survey metrics}, where condition was the fixed effect. 

%line for gaze analysis, freq of looking at the task

\section{Results}
\label{Sec:Results}
Eight participants were excluded from data analysis due to feedback and control issues that affected task performance. One participant mentioned that they could not feel the vibrotactile stimulus at all. The pressure sensor was not functioning for another participant, while the contact-location sensor was not functioning for a third participant. Finally, five participants experienced unreliable EMG signal quality throughout the experiment, as evidenced by their high average number of re-zeroing actions (at least two per trial). The following results represent the data from the remaining 40 participants; ten were in each of the four conditions.

\begin{figure}[tb!]
		\centering
		\vspace{1em}
		\includegraphics[width=.75\columnwidth]{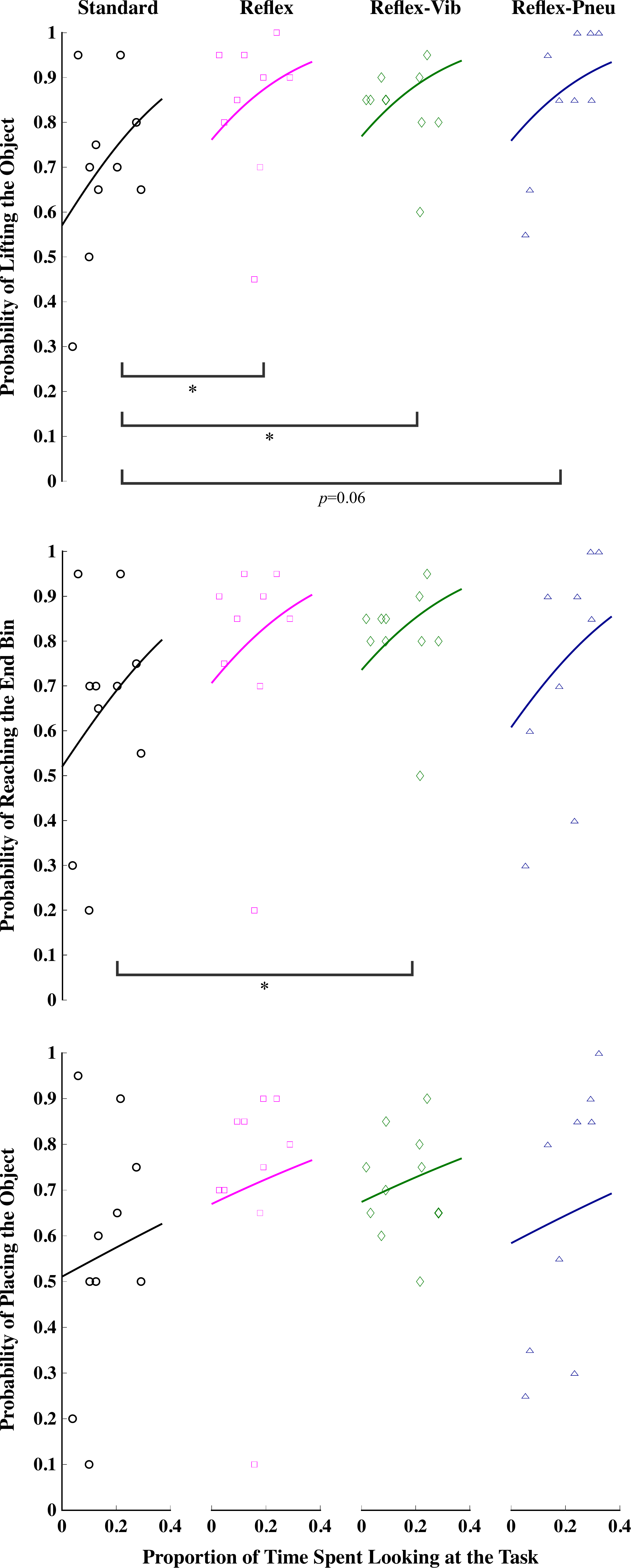}
	   % \vspace{-1.75em}
		\caption{The probability of accomplishing the three task milestones versus the proportion of time spent cheating, by condition. Solid lines indicate the average predicted probabilities from the mixed models, while individual markers show the average metric for each participant.}
		\label{fig:Milestones}
% 		\vspace{-1em}
\end{figure}

\begin{table*}[!t]
\vspace{-1em}
\caption{Summary of model statistics for odds of reaching task milestones}
\vspace{-1em}
\label{Table:TaskMilestones}
\tiny
\centering
\resizebox{\textwidth}{!}{%
\begin{tabular}{lccc|ccc|ccc|ccc|ccc|ccc} %{l c c c c c c c c c c c c c c c c}
\hline

& \multicolumn{3}{c}{Intercept (Standard)} & \multicolumn{3}{c}{Reflex} &  \multicolumn{3}{c}{Reflex-Vib} & \multicolumn{3}{c}{Reflex-Pneu} & \multicolumn{3}{c}{Cheating} & 
\multicolumn{3}{c}{Trial Number}\\ 
\hline
& $\beta$ & SE & $p$ & $\beta$ & SE & $p$ & $\beta$ & SE & $p$ & $\beta$ & SE & $p$ & $\beta$ & SE & $p$ & $\beta$ & SE & $p$ \\
Lifting object & 0.10 & 0.41 & 0.81 & 0.96 & 0.46 & 0.04 & 1.00 & 0.46 & 0.03 & 0.94 & 0.49 & 0.06 & 4.31 & 1.33 & 0.001 & 0.02 & 0.02 & 0.18\\
Reaching bin & --0.26 & 0.46 & 0.58 & 0.94 & 0.55 & 0.09 & 1.11 & 0.55 & 0.04 & 0.42 & 0.56 & 0.45 & 4.22 & 1.29 & 0.001 & 0.04 & 0.02 & 0.04 \\
Placing object & --0.47 & 0.43 & 0.27 & 0.77 & 0.50 & 0.12 & 0.82 & 0.50 & 0.10 & 0.33 & 0.52 & 0.53 & 2.14 & 1.14 & 0.06 & 0.04 & 0.02 & 0.007\\
\hline
% \footnote[2]{} Median reported. \footnote[8]{} Interquartile range reported.
\end{tabular}}
\end{table*}

% \begin{table*}[!b]
% \vspace{-1em}
% \caption{Summary of model statistics for time to achieve milestones}
% \vspace{-1em}
% \label{Table:TimeMilestones}
% \tiny
% \centering
% \resizebox{\textwidth}{!}{%
% \begin{tabular}{lccc|ccc|ccc|ccc|ccc} %{l c c c c c c c c c c c c c c c c}
% \hline

% & \multicolumn{3}{c}{Intercept (Standard)} & \multicolumn{3}{c}{Reflex} &  \multicolumn{3}{c}{Reflex-Vib} & \multicolumn{3}{c}{Reflex-Pneu} & \multicolumn{3}{c}{Pr. Cheating}\\ 
% \hline
% & $\beta$ & SE & $p$ & $\beta$ & SE & $p$ & $\beta$ & SE & $p$ & $\beta$ & SE & $p$ & $\beta$ & SE & $p$\\
% Lifting object & 26.4 & 1.89 & $<$ 0.001 & -2.84 & 2.31 & 0.23 & -1.37 & 2.27 & 0.55 & -2.70 & 2.36 & 0.26 & 1.59 & 6.64 & 0.81\\
% Reaching bin & 27.9 & 2.15 & $<$ 0.001 & -3.30 & 2.65& 0.22 & -0.99 & 2.63 & 0.71 & -1.51 & 2.79 & 0.59 & 6.84 & 7.08 & 0.34\\
% Placing object & 30.1 & 2.08 & $<$ 0.001 & -1.57 & 2.51 & 0.54 & 1.12 & 2.49 & 0.64 & -1.07 & 2.67 & 0.69 & 13.9 & 7.18 & 0.05\\
% \hline
% % \footnote[2]{} Median reported. \footnote[8]{} Interquartile range reported.
% \end{tabular}}
% \end{table*}

\begin{table*}[!tb]
\vspace{-1em}
\caption{Summary of model statistics for time to achieve milestones}
\vspace{-1em}
\label{Table:TimeMilestones}
\tiny
\centering
\resizebox{\textwidth}{!}{%
\begin{tabular}{lccc|ccc|ccc|ccc|ccc|ccc} %{l c c c c c c c c c c c c c c c c}
\hline

& \multicolumn{3}{c}{Intercept (Standard)} & \multicolumn{3}{c}{Reflex} &  \multicolumn{3}{c}{Reflex-Vib} & \multicolumn{3}{c}{Reflex-Pneu} & \multicolumn{3}{c}{Cheating} & \multicolumn{3}{c}{Trial Number}\\ 
\hline
& $\beta$ & SE & $p$ & $\beta$ & SE & $p$ & $\beta$ & SE & $p$ & $\beta$ & SE & $p$ & $\beta$ & SE & $p$ & $\beta$ & SE & $p$\\
Lifting object & 30.8 & 2.28 & $<$ 0.001 & -3.26 & 2.29 & 0.16 & -1.53 & 2.24 & 0.50 & -2.47 & 2.34 & 0.30 & -3.57 & 6.76 & 0.60 & -0.35 & 0.11 & $<$0.001\\
Reaching bin & 33.0 & 2.59 & $<$ 0.001 & -3.68 & 2.68 & 0.18 & -1.41 & 2.66 & 0.60 & -1.20 & 2.82 & 0.67 & 0.63 & 7.25 & 0.93 & -0.39 & 0.11 & $<$ 0.001\\
Placing object & 36.8 & 2.59 & $<$ 0.001 & -1.82 & 2.54 & 0.48 & 0.71 & 2.53 & 0.78 & -0.52 & 2.70 & 0.85 & 5.59 & 7.36 & 0.45 & -0.50 & 0.11 & $<$ 0.001\\
\hline
% \footnote[2]{} Median reported. \footnote[8]{} Interquartile range reported.
\end{tabular}}
\end{table*}
\subsection{Task Milestones}
% The odds of lifting the object from the start bin was not significantly different from chance in the Standard condition. 
% % ($\beta$~=~0.39, $SE$~=~0.35, $p$~=~0.26). 
% However, Reflex 
% % ($\beta$~=~0.95, $SE$~=~0.46, $p$~=~0.04)
% and Reflex-Vib 
% % ($\beta$~=~0.98, $SE$~=~0.46, $p$~=~0.03) 
% significantly improved those odds compared to the Standard. The effect of Reflex-Pneu was approaching significance over the Standard condition. 
% % ($\beta$~=~0.94, $SE$~=~0.49, $p$~=~0.055). 
% An increased frequency of looking away significantly increased the odds of lifting the object.The number of trials had no effect on the odds of lifting the object. 
% % ($\beta$ = 3.99, $SE$ = 1.30, $p$~=~0.002). 
% % Thus, when controlling for the frequency of looking awa     y and the trial number, both Reflex and Reflex-Vib significantly increased the odds of being able to lift the object, in comparison with the Standard condition.

%%% Jeremy's restructure of previous paragraph
Fig. \ref{fig:Milestones} and Table \ref{Table:TaskMilestones} show complete task milestones results. An increased amount of cheating significantly increased the participant's odds of lifting the object. The trial number, however, had no effect on the odds of lifting the object. When controlling for amount of cheating and number of trials, both Reflex and Reflex-Vib significantly increased the odds of being able to lift the object, in comparison with the Standard condition. The same comparison is close to significant ($p$~=~0.06) for Reflex-Pneu. The odds of lifting the object from the start bin in the Standard condition was not significantly different from 50\%.
% (see Fig. \ref{fig:Milestones} and Table \ref{Table:TaskMilestones} for complete results).

% The odds of reaching the end bin was not significantly different from chance in the Standard condition. 
% % ($\beta$~=~0.19, $SE$~=~0.40, $p$~=~0.64). 
% However, Reflex-Vib 
% % ($\beta$~=1.08, $SE$~=~0.55, $p$~=~0.047) 
% significantly improved those odds compared to the standard condition. Contrary to this, Reflex
% % ($\beta$~=~0.93, $SE$~=~0.54, $p$~=~0.086) 
% and Reflex-Pneu 
% % ($\beta$~=~0.42, $SE$~=~0.56, $p$~=~0.45) 
% were not significantly better. An increased frequency of looking away significantly improved the odds of reaching the bin. Similarly, more trials significantly increased the odds of reaching the bin.
% % ($\beta$ = 3.67, $SE$ = 1.26, $p$~=~0.004).
% Thus, when controlling for the frequency of looking away and trial number, only Reflex-Vib significantly increased the odds of being able to move the object to the end bin, in comparison with the standard condition.

%%% Jeremy's restructure of previous paragraph
An increased amount of cheating also significantly improved the odds of reaching the bin. Similarly, higher trial number (experience with the task) significantly increased the odds of reaching the bin. When controlling for amount of cheating and number of trials, only Reflex-Vib significantly increased the odds of being able to move the object to the end bin, in comparison with the Standard condition. The odds of reaching the end bin in the Standard condition was not significantly different from 50\%. 
% (see Fig. \ref{fig:Milestones} and Table \ref{Table:TaskMilestones} for complete results).

% The odds of setting the object into the end bin was not different than chance in the Standard condition 
% % ($\beta$~=~0.06, $SE$~=~0.38, $p$~=~0.87).
% Reflex 
% % ($\beta$~=~0.76, $SE$~=~0.5, $p$~=~0.13), 
% Reflex-Vib 
% % ($\beta$~=~0.79, $SE$~=~0.50, $p$~=~0.12), 
% and Reflex-Pneu 
% % ($\beta$~=~0.34, $SE$~=~0.52, $p$~=~0.51) 
% were no better than the Standard condition. An increased frequency of looking away from the visual target did not affect the odds of complete success. However, more practice significantly improved the odds of complete success. 
% % ($\beta$~=~1.47, $SE$~=~1.12, $p$~=~0.19). 
% All conditions had a similar effect on the odds of complete success in the task, regardless of the amount of time spent cheating or the trial number. Detailed statistics can be found in Table \ref{Table:TaskMilestones}. 

%%% Jeremy's restructure of previous paragraph
An increased amount of cheating (looking toward the task) was close to significant in affecting the odds of setting the object in the end bin ($p=0.06$). However, higher trial number significantly improved the odds of complete success. When controlling for the amount of cheating and trial number, no condition resulted in odds that were significantly better than 50\%.

% A depiction of the effect of the frequency of looking away from the visual target on the outcomes for each task and condition is shown in Fig. \ref{fig:Milestones}.

\begin{figure}[t]
		\centering
		\vspace{1em}
		\includegraphics[width=\columnwidth]{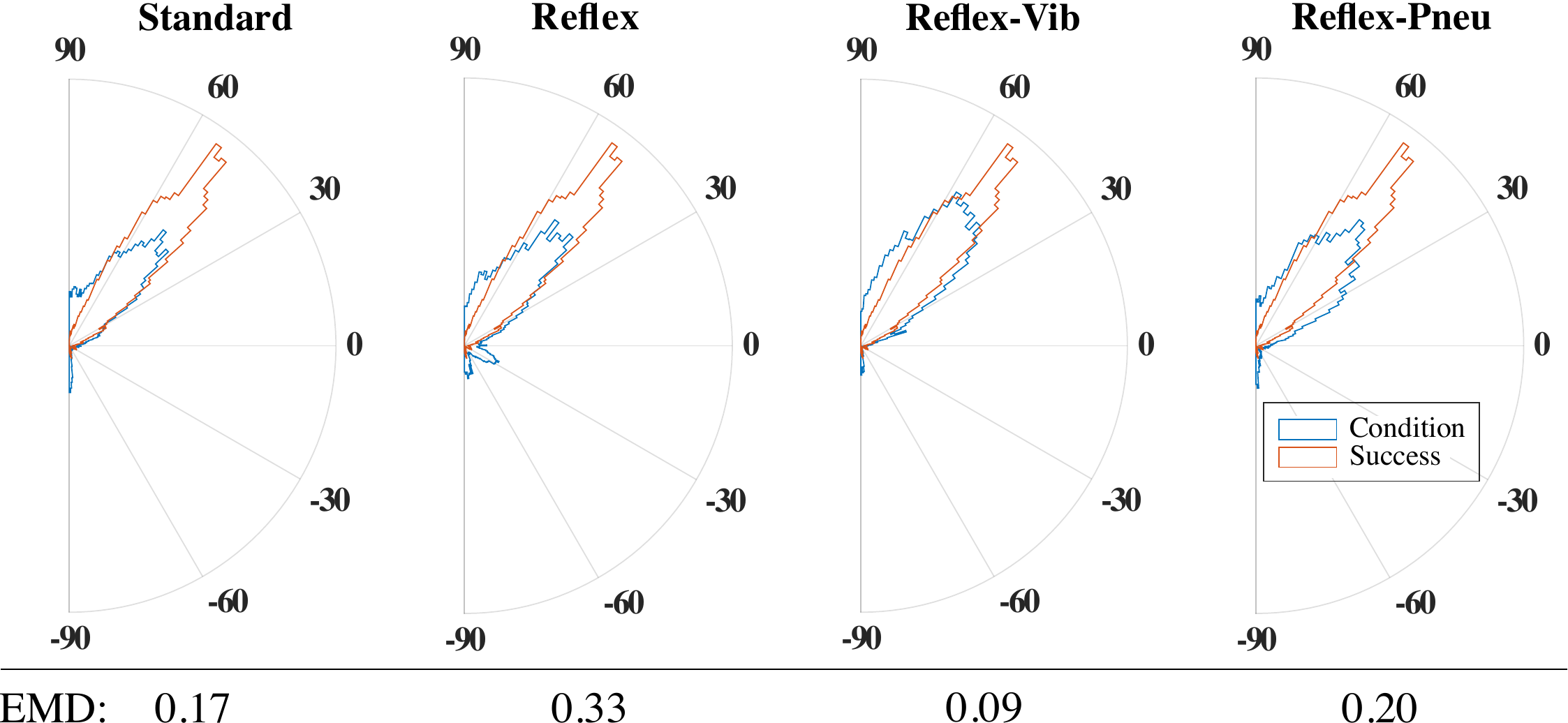}
	   % \vspace{-1.75em}
		\caption{Normalized polar histograms for the relative angle (degrees) between the prosthesis fingertips and the object. The Earth Mover's Distance (EMD) was computed between the successful grasp histogram and each grasp histograms from all conditions. A smaller value indicates more similarity with the successful grasp histogram.}
		\label{Fig:grasphistogram}
		\vspace{-1em}
\end{figure}

\begin{table*}[!t]
\vspace{-1em}
\caption{Summary of model statistics for survey results}
\vspace{-1em}
\label{Table:Survey}
\tiny
\centering
\resizebox{\textwidth}{!}{%
\begin{tabular}{lccc|ccc|ccc|ccc} %{l c c c c c c c c c c c c c c c c}
\hline

& \multicolumn{3}{c}{Intercept (Standard)} & \multicolumn{3}{c}{Reflex} &  \multicolumn{3}{c}{Reflex-Vib} & \multicolumn{3}{c}{Reflex-Pneu} \\ 
\hline
& $\beta$ & SE & $p$ & $\beta$ & SE & $p$ & $\beta$ & SE & $p$ & $\beta$ & SE & $p$\\
Finding object & 70.8 & 6.97 & $<$ 0.001 & 1.1 & 9.86 & 0.91 & 4.7 & 9.86 & 0.64 & --7.1 & 9.86 & 0.48\\
Grasping object & 41.1 & 6.51 & $<$ 0.001 & 7.1 & 9.21 & 0.52 & --6.0 & 9.21 & 0.52 & 4.1 & 9.21 & 0.66\\
Lifting object & 75.4 & 6.52 & $<$ 0.001 & --2.8 & 9.23 & 0.76 & --7.7 & 9.23 & 0.41 & --13.6 & 9.23 & 0.15\\
Moving object & 67.1 & 6.66 & $<$ 0.001 & 2.9 & 9.42 & 0.76 & 7.6 & 9.42 & 0.43 & --5.7 & 9.42 & 0.55\\
Placing object & 75.4 & 5.93 & $<$ 0.001 & --21.3 & 8.39 & 0.02 & --5.1 & 8.39 & 0.55 & --8.4 & 8.39 & 0.32\\
Mental effort & 60.9 & 6.73 & $<$ 0.001 & --1.0 & 9.53 & 0.92 & 3.1 & 9.53 & 0.75 & --1.5 & 9.53 & 0.88\\
Physical effort & 49.6 & 6.49 & $<$ 0.001 & --3.5 & 9.18 & 0.71 & 9.4 & 9.18 & 0.31 & 14.8 & 9.18 & 0.12\\
Physical comfort & 64.5 & 6.67 & $<$ 0.001 & --1.9 & 9.43 & 0.84 & --3.7 & 9.43 & 0.70 & --14.3 & 9.43 & 0.14\\
Frustration & 50.9 & 6.32 & $<$ 0.001 & --4.4 & 8.94 & 0.63 & 9.40 & 8.94 & 0.30 & --1.9 & 8.94 & 0.83\\
Time pressure & 59.6 & 7.27 & $<$ 0.001 & --14.7 & 10.3 & 0.16 & --12.2 & 10.3 & 0.24 & --10.3 & 10.3 & 0.32\\
Auditory cues & 51.6 & 8.52 & $<$ 0.001 & 4.4 & 12.1 & 0.71 & 3.9 & 12.1 & 0.75 & 0.1 & 12.1 & 0.99\\
Visual cues & 34.3 & 7.93 & $<$ 0.001 & --1.0 & 11.1 & 0.92 & 7.5 & 11.1 & 0.50 & 19.1 & 11.1 & 0.09\\
Somatosensory cues & 85.2 & 4.22 & $<$ 0.001 & --2.9 & 5.96 & 0.63 & 2.5 & 5.96 & 0.68 & --11.8 & 5.96 & 0.055\\
\hline
% \footnote[2]{} Median reported. \footnote[8]{} Interquartile range reported.
\end{tabular}}
\end{table*}

Table \ref{Table:TimeMilestones} displays detailed statistics for milestone timing. An increased amount of cheating did not significantly affect the time required to complete any of the task milestones. The time required to reach each of the milestones did not significantly differ by condition. However, the trial number did significantly decrease the time required to lift, move, and set the object down, again showing the benefits of task experience.

% \begin{table*}[!b]
% \vspace{-1em}
% \caption{Summary of model statistics for interaction quality}
% \vspace{-1em}
% \label{Table:InteractionQuality}
% \tiny
% \centering
% \resizebox{\textwidth}{!}{%
% \begin{tabular}{lccc|ccc|ccc|ccc|ccc|ccc} %{l c c c c c c c c c c c c c c c c}
% \hline

% & \multicolumn{3}{c}{Intercept (Standard)} & \multicolumn{3}{c}{Reflex} &  \multicolumn{3}{c}{Reflex-Vib} & \multicolumn{3}{c}{Reflex-Pneu} & \multicolumn{3}{c}{Pr. Cheating} & \multicolumn{3}{c}{Success}\\ 
% \hline
% & $\beta$ & SE & $p$ & $\beta$ & SE & $p$ & $\beta$ & SE & $p$ & $\beta$ & SE & $p$ & $\beta$ & SE & $p$  $\beta$ & SE & $p$\\
% Drops & 0.47 & 0.11 & $<$ 0.001 & 0.28 & 0.11 & 0.02 & 0.13 & 0.12 & 0.25 & 0.18 & 0.12 & 0.14 & 0.70 & 0.31 & 0.02 & --0.48 & 0.07 & $<$ 0.001\\
% Integral Lateral Forces & 993.8 & 204.0 & $<$ 0.001 & 440.33 & 284.23 & 0.13 & 304.9 & 284.2 & 0.29 & 303.3 & 292.2 & 0.31 & 163.4 & 221.8 & 0.46 & --359.1 & 39.1 & $<$ 0.001\\
% Integral Normal Forces & 1253.7 & 111.2 & $<$ 0.001 & 227.1 & 152.7 & 0.15 & 147.6 & 152.7 & 0.34 & 81.1 & 157.1 & 0.61 & 166.5 & 166.6 & 0.32 & --334.2 & 29.6 & $<$ 0.001\\
% \hline
% % \footnote[2]{} Median reported. \footnote[8]{} Interquartile range reported.
% \end{tabular}}
% \end{table*}

\subsection{Number of Drops}
Trial number had no effect on the number of drops ($\beta$~=~\mbox{--0.004}, $SE$~=~0.004, $p$~=~0.33). The number of drops in the Standard condition was significantly greater than 0 ($\beta$~=~0.22, $SE$~=~0.09, $p$~$<$~0.02).  Both the Reflex and the Reflex-Pneu conditions caused a significant increase in the number of drops (Reflex: $\beta$~=~0.24, $SE$~=~0.11, $p$~=~0.046; Reflex-Pneu: $\beta$~=~0.23, $SE$~=~0.11, $p$~=~0.049). The number of drops in the Reflex-Vib condition did not differ from that in the Standard condition ($\beta$~=~0.12, $SE$~=~0.11, $p$~=~0.30).

\subsection{Grasping Location}
Fig. \ref{Fig:grasphistogram} shows the polar histogram of the relative angles between the prosthetic finger and the object during attempted grasping for each condition compared to all successful grasps. Successful grasps are most often found between 40 and 60 degrees. The odds of being within the optimal grasp angle range were significantly less than 50\% in the Standard condition ($\beta$~=~--0.45, $SE$~=~0.10, $p$~$<$~0.001). Reflex and Reflex-Pneu had significantly lower odds compared to the Standard condition (Reflex: $\beta$~=~--0.25, $SE$~=~0.12, $p$~=~0.038; Reflex-Pneu: $\beta$~=~--0.31, $SE$~=~0.10, $p$~=~0.002). However, Reflex-Vib did not significantly differ from the Standard condition ($\beta$~=~0.07, $SE$~=~0.11, $p$~=~0.55). Post-hoc tests with a Bonferroni correction indicated that participants in Reflex-Vib had significantly better odds of being in the successful grasping range than those in the Reflex ($\beta$~=~0.32, $SE$~=~0.14, $p$~=~0.039) and Reflex-Pneu ($\beta$~=~0.38, $SE$~=~0.13, $p$~=~0.009) conditions. The earth mover's distance (EMD) metrics calculated for each condition support the results of the mixed-model analysis. 

\subsection{Cheating Frequency}
The frequency of looking away from the visual target was greater than zero in the Standard condition ($\beta$~=~0.15, $SE$~=~0.03, $p$~$<$~0.001). The Reflex ($\beta$~=~--0.001, $SE$~=~0.04, $p$~=~0.97), Reflex-Vib ($\beta$~=~--0.007, $SE$~=~0.04, $p$~=~0.86), and Reflex-Pneu ($\beta$~=~0.05, $SE$~=~0.04, $p$~=~0.24) conditions did not significantly differ from the Standard condition. Thus, there was an equal amount of cheating across all conditions.

\subsection{Survey}
Participants in the Standard condition provided ratings for all survey questions that were significantly different from 0 (see Table \ref{Table:Survey} for complete results). The majority of survey responses did not significantly differ by condition, except for the following few cases. Participants in the Reflex condition rated their ability to set the object down (placing object) as significantly lower than the Standard condition. In a post-hoc test with a Bonferroni correction, participants in the Reflex-Pneu condition rated their use of somatosensory cues as significantly lower than those in the Reflex-Vib condition ($\beta$~=~--14.3, $SE$~=~5.96, $p$~=~0.02).

\section{Discussion}
In this study, we investigated how autonomous reflexes and two different forms of haptic feedback affect performance in a reach-to-pick-and-place task using a myoelectric upper-limb prosthesis without direct visual feedback. Our intent was to replicate tasks where observation is undesirable or impossible. We compared four conditions in a between-subjects study: a standard prosthesis, a prosthesis with reflex controllers to mitigate object slip and excessive grasping (Reflex condition), a prosthesis with reflex controllers and vibrotactile feedback of contact location (Reflex-Vib condition), and a prosthesis with reflex controllers and spatial pressure-based feedback of contact location (Reflex-Pneu condition). We also presented the design and characterization of a novel contact-location sensor that enabled the reflex controllers and haptic feedback. 

While the prosthesis with reflex controllers improved the odds of lifting the object compared to the standard prosthesis condition, the prosthesis with reflex controllers and vibrotactile feedback was the only condition to improve performance in both lifting and moving the object to the end bin. The reflex controller that contributed the most to these results was likely the anti-overgrasping controller, as this system was active for every grasp attempt. Contrarily, the slip prevention algorithms were not always triggered for every grasp attempt. The fast slip controller was active approximately three times per trial, while the slow slip prevention was active about two times per trial; this indicates that fast slips were the more common type of slip. Compared to reflexes alone, the vibration feedback improved grasping location accuracy, enabling a more secure grasp that was more robust to disturbances introduced during the transportation phase. This result aligns with previous research, where vibration feedback was shown to be especially relevant for grasp-and-lift tasks, enabling grasp consistency during fragile object manipulation \cite{Engels2019WhenHand}. That no condition outperformed the standard prosthesis in being able to place the object in the end bin is likely due to the fact that the most difficult parts of the task are the lifting and moving stages. Once a participant accomplishes the first two milestones, it is straightforward to place the object in the end bin given enough practice, regardless of the condition. Indeed, only the trial number, which represents task experience, played a significant role in the outcome of placing the object in the end bin.

% vib and Reflex improved lifting. only vib improved reaching bin. vib improved grasp location. Reflex more drops. indicates that automatic slip prevention/overgrasp is not enough, vibrotactile feedback provides another advantage. vibration allows for more correct placement

% why not success? -> participant mentioned difficulty in setting down (finding the second bin), some people ran out of time, others set the object just outside the bin, object fell out when banging edges of bin. Once the object has been lifted or bin has been reached, there is no difference between conditions in setting the object down into the bin.

Contrary to expectations, the modality-matched haptic feedback in the Reflex-Pneu condition resulted in similar performance to the Standard prosthesis condition, indicating that the benefits of the reflex controller were cancelled out by the pneumatic pressure feedback. 
% as evidenced by lack of significant
% It would seem that any type of haptic feedback would not work equally well for this task, as demonstrated by the lack of significant improvement in the Reflex-Pneu condition for any of the milestones.  
Normally, modality-matched haptic feedback is thought to be easier to understand than non-matched feedback \cite{Kim2010OnProsthetics, Stephens-Fripp2018} and has been perceived favorably by amputees \cite{Wijk2020SensoryUse}. However, previous work has also shown no functional benefit of more modality-matched haptic feedback compared to a non-matched modality \cite{Thomas2019}. 
% In that study, we ensured that each haptic cue was perceived the same for each participant using the cross-modal matching approach first presented in \cite{Pitts2014}. 
Therefore, we postulate that the difference between the vibrotactile and pressure feedback in our study stems from the discriminability of the way feedback was presented: amplitude discrimination of a single tactor versus spatial discrimination of eight bellows. Based on this finding, it seems that not all types of haptic feedback are equal, and some may provide no quantifiable improvement over simpler alternatives. 

The Bellowband's pneumatic bellows are 16\,mm in diameter and are spaced 24\,mm apart \cite{Young2019Bellowband:Vibration}. For the age range tested, this is slightly below the spatial discrimination of 30\,mm on the back of the arm \cite{Stevens2009SpatialSpan}. Prior work \cite{Guemann2019EffectAmputees} also reports that the localization accuracy for a circular array of 6 vibrotactile motors equally spaced at 30\,mm around the upper arm was around 42\%. While this result is specifically for 20\,mm diameter vibrotactors, we argue that the pulsed activation of the bellows is similar to vibration feedback. Furthermore, this value likely represents the upper bound for localization accuracy in the Reflex-Pneu condition due to the smaller spacing between bellows and the added difficulty of performing the reach-to-pick-and-place task without direct vision. Moreover, only four of the eight bellows represented the critical region of the inner part of the prosthesis finger. This is in contrast to the vibrotactile feedback, which has eight discriminable amplitude levels based on a difference threshold of 0.3 \cite{Choi2013VibrotactileApplications}. So although the Reflex-Pneu feedback condition was modality-matched with respect to feedback location mapping and feedback type (i.e., pressure), it likely suffered from lower resolution compared to the vibrotactile condition. Survey feedback supports this idea: participants in the Reflex-Pneu condition felt that they used somatosensory cues less than those in the Reflex-Vib condition, indicating a breakdown in the understanding of the localized pressure feedback compared to vibrotactile feedback.

Although one could alter the original design presented in \cite{Young2019Bellowband:Vibration} by changing the spacing between the bellows, this change would also reduce the total number of bellows and consequently the overall resolution of the haptic feedback. Future work to improve the Bellowband includes reducing the size of the bellows themselves so that more bellows can be added while maintaining an appropriate spacing. 
% other types of ineffective haptic feedback?

Due to the challenges of interpreting and using unfamiliar feedback, participants in both Reflex-Vib and Reflex-Pneu conditions may have seen improvement in performance with extended training on their particular feedback modalities. This additional training may also improve grasping location accuracy over the Standard condition. Several comments in the surveys from both conditions confirmed this uncertainty in using the feedback to precisely determine the correct grasping location. Previous literature has also indicated that more practice with haptic feedback yields substantial benefits \cite{Stepp2012RepeatedPerformance}. 
Another possible improvement for vibrotactile feedback in this study would be to customize the mapping function of the sensor signal to the vibration amplitude. The current mapping (Eq. \ref{Eqn:VibMapping}) may not have maximized differences within the critical inner region of the contact-location sensor, as demonstrated in Fig. \ref{fig:characterization}. Furthermore, optimizing the sensor construction for the curved surfaces of the fingers would improve the sensor's activation profile and thus facilitate improved feedback strategies. Finally, future work should include psychometric and psychophysical assessments of the feedback modalities to ensure the feedback works as intended \cite{Marasco2011,Shehata2018,DAnna2019}. 

% Discussion on grasping/obj drops/strategy
Participants in both the Standard and Reflex-Vib conditions were more likely to correctly orient the prosthetic hand for optimal grasping, while grasping in the Reflex and Reflex-Pneu conditions were the most dissimilar. This could indicate that participants in the Reflex condition employed a suboptimal grasping strategy. The over-grasp reflex controller caused the prosthesis to stop closing once an object made contact with the thumb. This could have encouraged participants to make fast and frequent grasp attempts without much consideration for finger placement. This hypothesis is supported by the observed higher number of drops with the reflex controller compared to the Standard condition, and the significantly lower ratings of participants' ability to set the object in the bin with just the synthetic reflexes. On the other hand, participants in the Standard condition likely realized that fast grasp attempts would cause the object to slip out of the prosthesis' grasp. They were thus incentivized to find an ideal grasping position, which may have made them more aware of incidental mechanical cues transmitted through the prosthesis to their arm. In fact, two participants in the Standard condition remarked that the mechanical sensation of the contact-location sensor touching the object helped them orient the hand relative to the object. Previous research has indicated that incidental feedback is adequate to appropriately tune grasping force levels \cite{Markovic2018}. However, this type of feedback in which mechanical impacts are transmitted through the prosthesis to the user may be dampened if the prosthetic hand is encased in a rubber aesthetic glove. Nevertheless, future research could investigate ways to maximize the discriminability of incidental feedback by customizing fingerpads with ridges, bumps or other mechanical features to assist with localization.

Although there were no statistical differences between the Standard and Reflex-Vib feedback conditions in terms of successful grasping positions, participants in the Reflex-Vib condition still achieved higher performance in lifting and moving the object. Thus, in addition to correctly positioning the prosthesis for grasping, Reflex-Vib participants must have also appropriately modulated their grasping force, likely aided by the synthetic reflexes. Furthermore, participants in the Reflex-Vib condition positioned the prosthesis more accurately than participants in the Reflex and Reflex-Pneu conditions, indicating that reflex controllers alone are not enough to fully optimize performance, and higher resolution haptic feedback is needed. All things considered, without the combination of effective haptic feedback and reflex controllers, the studied task is difficult to perform without direct visual observation. This lack of tactile feedback and control is analogous to how performance deteriorates in reach-to-grasp tasks performed with anesthetized fingers \cite{Gentilucci1997TactileMovements} or by deafferented patients \cite{Parry2021AnticipationNeuropathy,Carteron2016TemporaryTasks,Jeannerod1984THELESION}. Although the experimental task here is closely related to activities of daily living (ADL), future work should also include established tests to evaluate the utility of the system more directly in ADL \cite{Hill2009FunctionalGroup}. Improvements to the grasping controller include allowing the user to override the control to produce larger grip forces that may be required for heavier objects. If the presented results can be validated with amputee users and different types of objects, we believe future myoelectric upper-limb prostheses should include tactile sensing, automatic reflexes, and socket-integrated vibrotactile feedback about contact. In addition, these results can be extended to even multi-grasp myoelectric hands; as in the present study, knowledge of contact location can be used to adjust the prosthesis location for grasping without direct vision. Furthermore, it could even help inform users as to which grasp type is most appropriate given the initial contact point. More broadly, the findings presented here could be used to improve other teleoperated systems such as robotic surgery.

% \section{Conclusion}
% In this paper, we investigated the utility of reflex controllers and haptic feedback (separately and combined) for non-amputee users of a standard myoelectric prosthesis equipped with custom tactile sensors that measure contact location and pressure. These conditions were evaluated in a reach-to-pick-and-place task conducted without direct vision. Results indicate that vibrotactile feedback combined with reflex controllers was the only condition to outperform the standard myoelectric prosthesis in achieving two of the three task milestones. Likewise, the single amplitude-modulated and frequency-modulated vibrotactor displayed task-relevant contact location more effectively that eight pneumatic tactors. These results can help inform improvements to smart prostheses as well as the broader fields of teleoperation and robotic manipulation.

\section*{Acknowledgment}
The authors thank Eric M.\ Young for providing the Bellowband. We also thank the US-German Fulbright Program, the Germanistic Society of America, Mastercard, the National Science Foundation Graduate Fellowship, and the Max Planck Society for funding the first author. Finally, we thank the International Max Planck Research School for Intelligent Systems (IMPRS-IS) for supporting Farimah Fazlollahi.
\vspace{-.1em}
\ifCLASSOPTIONcaptionsoff
  \newpage
\fi



%\bibliographystyle{IEEEtran}


% \end{thebibliography}
\bibliographystyle{IEEEtranNoURL}
\bibliography{references.bib}

\begin{thebibliography}{10}
\providecommand{\url}[1]{#1}
\csname url@rmstyle\endcsname
\providecommand{\newblock}{\relax}
\providecommand{\bibinfo}[2]{#2}
\providecommand\BIBentrySTDinterwordspacing{\spaceskip=0pt\relax}
\providecommand\BIBentryALTinterwordstretchfactor{4}
\providecommand\BIBentryALTinterwordspacing{\spaceskip=\fontdimen2\font plus
\BIBentryALTinterwordstretchfactor\fontdimen3\font minus
  \fontdimen4\font\relax}
\providecommand\BIBforeignlanguage[2]{{%
\expandafter\ifx\csname l@#1\endcsname\relax
\typeout{** WARNING: IEEEtran.bst: No hyphenation pattern has been}%
\typeout{** loaded for the language `#1'. Using the pattern for}%
\typeout{** the default language instead.}%
\else
\language=\csname l@#1\endcsname
\fi
#2}}

\bibitem{Karl2013NonvisualReach}
\BIBentryALTinterwordspacing
J.~M. Karl, L.~R. Schneider, and I.~Q. Whishaw, ``{Nonvisual Learning of
  Intrinsic Object Properties in a Reaching Task Dissociates Grasp from
  Reach},'' \emph{Exp Brain Res}, vol. 225, pp. 465--477, 2013.
\BIBentrySTDinterwordspacing

\bibitem{Winges2003TheGrasp}
S.~A. Winges, D.~J. Weber, and M.~Santello, ``{The role of vision on hand
  preshaping during reach to grasp},'' \emph{Exp Brain Res}, vol. 152, pp.
  489--498, 2003.

\bibitem{Gentilucci1997TactileMovements}
M.~Gentilucci, I.~Toni, E.~Daprati, and M.~Gangitano, ``{Tactile Input of the
  Hand and the Control of Reaching to Grasp Movements},'' \emph{Experimental
  Brain Research}, vol. 114, pp. 130--137, 1997.

\bibitem{Johansson1996SensoryHumans}
R.~S. Johansson, ``{Sensory Control of Dexterous Manipulation in Humans},''
  \emph{Hand and Brain}, pp. 381--414, 1996.

\bibitem{Johansson1992Sensory-MotorActions}
R.~S. Johansson and K.~J. Cole, ``{Sensory-Motor Coordination During Grasping
  and Manipulative Actions},'' \emph{Current Opinion in Neurobiology}, vol.~2,
  pp. 815--823, 1992.

\bibitem{Nakajima2006Location-specificMuscles}
T.~Nakajima, M.~Sakamoto, T.~Endoh, and T.~Komiyama, ``{Location-specific and
  Task-dependent Modulation of Cutaneous Reflexes in Intrinsic Human Hand
  Muscles},'' \emph{Clinical Neurophysiology}, vol. 117, pp. 420--429, 2006.

\bibitem{Atkins1996EpidemiologicPriorities}
D.~J. Atkins, D.~C. Heard, and W.~H. Donovan, ``{Epidemiologic Overview of
  Individuals with Upper-limb Loss and Their Reported Research Priorities},''
  \emph{Journal of Prosthetics and Orthotics}, vol.~8, pp. 2--11, 1996.

\bibitem{Sobuh2014}
\BIBentryALTinterwordspacing
M.~M.~D. Sobuh, L.~P.~J. Kenney, A.~J. Galpin, S.~B. Thies, J.~McLaughlin,
  J.~Kulkarni, and P.~Kyberd, ``{Visuomotor Behaviours When Using a Myoelectric
  Prosthesis},'' \emph{Journal of NeuroEngineering and Rehabilitation},
  vol.~11, p.~72, 2014.
\BIBentrySTDinterwordspacing

\bibitem{Thomas2020}
\BIBentryALTinterwordspacing
N.~Thomas, G.~Ung, H.~Ayaz, and J.~D. Brown, ``{Neurophysiological Evaluation
  of Haptic Feedback for Myoelectric Prostheses},'' \emph{IEEE Transactions on
  Human-Machine Systems}, vol.~51, pp. 253--264, 2021.
\BIBentrySTDinterwordspacing

\bibitem{Ponraj2019ActiveGrippers}
\BIBentryALTinterwordspacing
G.~Ponraj, A.~V. Prituja, C.~Li, A.~Bamotra, Z.~Guoniu, S.~K. Kirthika, N.~V.
  Thakor, A.~B. Soares, and H.~L. Ren, ``{Active Contact Enhancements With
  Stretchable Soft Layers and Piezoresistive Tactile Array for Robotic
  Grippers},'' in \emph{Proc. 15th International Conference on Automation
  Science and Engineering}, 2019, pp. 1808--1813.
\BIBentrySTDinterwordspacing

\bibitem{Osborn2014}
\BIBentryALTinterwordspacing
L.~Osborn, W.~W. Lee, R.~Kaliki, and N.~Thakor, ``{Tactile Feedback in Upper
  Limb Prosthetic Devices Using Flexible Textile Force Sensors},'' in
  \emph{Proc. IEEE RAS/EMBS International Conference on Biomedical Robotics and
  Biomechatronics}, 8 2014, pp. 114--119.
\BIBentrySTDinterwordspacing

\bibitem{Lee2021Piezo}
\BIBentryALTinterwordspacing
H.~Lee, K.~Park, J.~Kim, and K.~J. Kuchenbecker, ``{Piezoresistive Textile
  Layer and Distributed Electrode Structure for Soft Whole-body Tactile
  Skin},'' \emph{Smart Materials and Structures}, vol.~30, p. 085036, 2021.
\BIBentrySTDinterwordspacing

\bibitem{Jimenez2014}
M.~C. Jimenez and J.~A. Fishel, ``{Evaluation of Force, Vibration and Thermal
  Tactile Feedback in Prosthetic Limbs},'' in \emph{Proc. IEEE Haptics
  Symposium (HAPTICS)}, 2014, pp. 437--441.

\bibitem{Antfolk2013b}
C.~Antfolk, C.~Cipriani, M.~C. Carrozza, C.~Balkenius, A.~Bj{\"{o}}rkman,
  G.~Lundborg, B.~Ros{\'{e}}n, and F.~Sebelius, ``{Transfer of Tactile Input
  from an Artificial Hand to the Forearm: Experiments in Amputees and
  Able-bodied Volunteers},'' \emph{Disability and Rehabilitation: Assistive
  Technology}, vol.~8, pp. 249--254, 2013.

\bibitem{Hartmann2014TowardsEmbodiment}
C.~Hartmann, J.~Linde, S.~Dosen, D.~Farina, L.~Seminara, L.~Pinna, M.~Valle,
  and M.~Capurro, ``{Towards Prosthetic Systems Providing Comprehensive Tactile
  feedback for Utility and Embodiment},'' \emph{Proc. IEEE Biomedical Circuits
  and Systems Conference, BioCAS}, pp. 620--623, 2014.

\bibitem{Shehata2020MechanotactileUse}
A.~W. Shehata, M.~Rehani, Z.~E. Jassat, and J.~S. Hebert, ``{Mechanotactile
  Sensory Feedback Improves Embodiment of a Prosthetic Hand During Active
  Use},'' \emph{Frontiers in Neuroscience}, vol.~0, p. 263, 2020.

\bibitem{Antfolk2013a}
\BIBentryALTinterwordspacing
C.~Antfolk, M.~D'Alonzo, M.~Controzzi, G.~Lundborg, B.~Rosen, F.~Sebelius, and
  C.~Cipriani, ``{Artificial Redirection of Sensation From Prosthetic Fingers
  to the Phantom Hand Map on Transradial Amputees: Vibrotactile Versus
  Mechanotactile Sensory Feedback},'' \emph{IEEE Transactions on Neural Systems
  and Rehabilitation Engineering}, vol.~21, pp. 112--120, 2013.
\BIBentrySTDinterwordspacing

\bibitem{Ward2018Multi-channelHand}
\BIBentryALTinterwordspacing
K.~Ward and D.~Pamungkas, ``{Multi-channel Electro-tactile Feedback System for
  a Prosthetic Hand},'' \emph{Mechatronics and Machine Vision in Practice},
  vol.~3, pp. 181--193, 2018.
\BIBentrySTDinterwordspacing

\bibitem{Scott1980Sensory-feedbackControl}
R.~N. Scott, H.~Brittain, R.~R. Caldwell, B.~Cameron, and V.~A. Dunfield,
  ``{Sensory-feedback system compatible with myoelectric control},''
  \emph{Biol. Eng. {\&} Comput}, vol.~18, pp. 65--69, 1980.

\bibitem{Antfolk2013SensoryProsthetics}
\BIBentryALTinterwordspacing
C.~Antfolk, M.~D. Alonzo, B.~Ros{\'{e}}n, G.~Lundborg, F.~Sebelius, and
  C.~Cipriani, ``{Sensory Feedback in Upper Limb Prosthetics},'' \emph{Expert
  Review of Medical Devices}, vol.~10, pp. 45--54, 2013.
\BIBentrySTDinterwordspacing

\bibitem{Salisbury1967APrehension}
\BIBentryALTinterwordspacing
L.~L. Salisbury and A.~B. Colman, ``{A mechanical hand with automatic
  proportional control of prehension},'' \emph{Medical and biological
  engineering 1967 5:5}, vol.~5, pp. 505--511, 9 1967.
\BIBentrySTDinterwordspacing

\bibitem{Chappell1987ControlHand}
\BIBentryALTinterwordspacing
P.~H. Chappell, J.~M. Nightingale, P.~J. Kyberd, and M.~Barkhordar, ``{Control
  of a single degree of freedom artificial hand},'' \emph{Journal of biomedical
  engineering}, vol.~9, pp. 273--277, 1987.
\BIBentrySTDinterwordspacing

\bibitem{Nightingale1985MicroprocessorArm}
J.~M. Nightingale, ``{Microprocessor control of an artificial arm},''
  \emph{Journal of Microcomputer Applications}, vol.~8, pp. 167--173, 4 1985.

\bibitem{Osborn2016}
L.~Osborn, R.~R. Kaliki, A.~B. Soares, and N.~V. Thakor, ``{Neuromimetic
  Event-Based Detection for Closed-Loop Tactile Feedback Control of Upper Limb
  Prostheses},'' \emph{IEEE Transactions on Haptics}, vol.~9, pp. 196--206,
  2016.

\bibitem{Matulevich2013UtilityControl}
B.~Matulevich, G.~E. Loeb, and J.~A. Fishel, ``{Utility of Contact Detection
  Reflexes in Prosthetic Hand Control},'' \emph{Proc. IEEE International
  Conference on Intelligent Robots and Systems}, pp. 4741--4746, 2013.

\bibitem{OttobockSensorHand}
\BIBentryALTinterwordspacing
{Ottobock}, ``{SensorHand Speed}.'' [Online]. Available:
  \url{https://shop.ottobock.us/Prosthetics/Upper-Limb-Prosthetics/Myo-Hands-and-Components/Myo-Terminal-Devices/SensorHand-Speed/p/8E38~58}
\BIBentrySTDinterwordspacing

\bibitem{Thomas2021Sensorimotor-inspiredVision}
\BIBentryALTinterwordspacing
N.~Thomas, F.~Fazlollahi, J.~D. Brown, and K.~J. Kuchenbecker,
  ``{Sensorimotor-inspired Tactile Feedback and Control Improve Consistency of
  Prosthesis Manipulation in the Absence of Direct Vision},'' in \emph{Proc.
  IEEE/RSJ International Conference on Intelligent Robots and Systems}, 2021.
\BIBentrySTDinterwordspacing

\bibitem{Kim2010OnProsthetics}
K.~Kim, J.~E. Colgate, J.~J. Santos-Munn{\'{e}}, A.~Makhlin, and M.~A. Peshkin,
  ``{On the Design of Miniature Haptic Devices for Upper Extremity
  Prosthetics},'' \emph{IEEE/ASME Transactions on Mechatronics}, vol.~15, pp.
  27--39, 2010.

\bibitem{Young2019Bellowband:Vibration}
E.~M. Young, A.~H. Memar, P.~Agarwal, and N.~Colonnese, ``{Bellowband: A
  Pneumatic Wristband for Delivering Local Pressure and Vibration},'' in
  \emph{Proc. IEEE World Haptics Conference, WHC}, 2019.

\bibitem{Nolan1982Two-PointWomen}
\BIBentryALTinterwordspacing
M.~F. Nolan, ``{Two-Point Discrimination Assessment in the Upper Limb in Young
  Adult Men and Women},'' \emph{Physical Therapy}, vol.~62, pp. 965--969, 1982.
\BIBentrySTDinterwordspacing

\bibitem{Koo2016}
\BIBentryALTinterwordspacing
J.-P. Koo, S.-H. Kim, H.-J. An, O.-G. Moon, J.-H. Choi, Y.-D. Yun, J.-H. Park,
  and K.-O. Min, ``{Two-point Discrimination of the Upper Extremities of
  Healthy Koreans in Their 20’s},'' \emph{Journal of Physical Therapy
  Science}, vol.~28, p. 870, 2016.
\BIBentrySTDinterwordspacing

\bibitem{Kyranou2018CausesProstheses}
\BIBentryALTinterwordspacing
I.~Kyranou, S.~Vijayakumar, and M.~S. Erden, ``{Causes of performance
  degradation in non-invasive electromyographic pattern recognition in upper
  limb prostheses},'' \emph{Frontiers in Neurorobotics}, vol.~12, p.~58, 9
  2018.
\BIBentrySTDinterwordspacing

\bibitem{Engels2019WhenHand}
L.~F. Engels, A.~W. Shehata, E.~J. Scheme, J.~W. Sensinger, and C.~Cipriani,
  ``{When less is more-discrete tactile feedback dominates continuous audio
  biofeedback in the integrated percept while controlling a myoelectric
  prosthetic hand},'' \emph{Frontiers in Neuroscience}, vol.~13, p. 578, 2019.

\bibitem{Stephens-Fripp2018}
\BIBentryALTinterwordspacing
B.~Stephens-Fripp, G.~Alici, and R.~Mutlu, ``{A Review of Non-Invasive Sensory
  Feedback Methods for Transradial Prosthetic Hands},'' \emph{IEEE Access},
  vol.~6, pp. 6878--6899, 2018.
\BIBentrySTDinterwordspacing

\bibitem{Wijk2020SensoryUse}
\BIBentryALTinterwordspacing
U.~Wijk, I.~K. Carlsson, C.~Antfolk, A.~Bj{\"{o}}rkman, and B.~Ros{\'{e}}n,
  ``{Sensory Feedback in Hand Prostheses: A Prospective Study of Everyday
  Use},'' \emph{Frontiers in Neuroscience}, vol.~14, 2020.
\BIBentrySTDinterwordspacing

\bibitem{Thomas2019}
\BIBentryALTinterwordspacing
N.~Thomas, G.~Ung, C.~McGarvey, and J.~D. Brown, ``{Comparison of Vibrotactile
  and Joint-torque Feedback in a Myoelectric Upper-limb Prosthesis},''
  \emph{Journal of NeuroEngineering and Rehabilitation}, vol.~16, p.~70, 2019.
\BIBentrySTDinterwordspacing

\bibitem{Stevens2009SpatialSpan}
\BIBentryALTinterwordspacing
J.~C. Stevens and K.~K. Choo, ``{Spatial Acuity of the Body Surface over the
  Life Span},'' \emph{Somatosensory {\&} Motor Research}, 2009.
\BIBentrySTDinterwordspacing

\bibitem{Guemann2019EffectAmputees}
\BIBentryALTinterwordspacing
M.~Guemann, S.~Bouvier, C.~Halgand, F.~Paclet, L.~Borrini, D.~Ricard,
  E.~Lapeyre, D.~Cattaert, and A.~d. Rugy, ``{Effect of Vibration
  Characteristics and Vibror Arrangement on the Tactile Perception of the Upper
  Arm in Healthy Subjects and Upper Limb Amputees},'' \emph{Journal of
  NeuroEngineering and Rehabilitation}, vol.~16, p. 138, 2019.
\BIBentrySTDinterwordspacing

\bibitem{Choi2013VibrotactileApplications}
\BIBentryALTinterwordspacing
S.~Choi and K.~J. Kuchenbecker, ``{Vibrotactile Display: Perception,
  Technology, and Applications},'' \emph{Proceedings of the IEEE}, vol. 101,
  pp. 2093--2104, 2013.
\BIBentrySTDinterwordspacing

\bibitem{Stepp2012RepeatedPerformance}
\BIBentryALTinterwordspacing
C.~E. Stepp, Q.~An, and Y.~Matsuoka, ``{Repeated Training with Augmentative
  Vibrotactile Feedback Increases Object Manipulation Performance},''
  \emph{PLoS ONE}, vol.~7, p. e32743, 2012.
\BIBentrySTDinterwordspacing

\bibitem{Marasco2011}
\BIBentryALTinterwordspacing
P.~D. Marasco, K.~Kim, J.~E. Colgate, M.~A. Peshkin, and T.~A. Kuiken,
  ``{Robotic touch shifts perception of embodiment to a prosthesis in targeted
  reinnervation amputees},'' \emph{Brain}, vol. 134, pp. 747--758, 3 2011.
\BIBentrySTDinterwordspacing

\bibitem{Shehata2018}
\BIBentryALTinterwordspacing
A.~W. Shehata, L.~F. Engels, M.~Controzzi, C.~Cipriani, E.~J. Scheme, and J.~W.
  Sensinger, ``{Improving internal model strength and performance of prosthetic
  hands using augmented feedback},'' \emph{Journal of NeuroEngineering and
  Rehabilitation}, vol.~15, p.~70, 2018.
\BIBentrySTDinterwordspacing

\bibitem{DAnna2019}
E.~D'Anna, G.~Valle, A.~Mazzoni, I.~Strauss, F.~Iberite, J.~Patton, F.~M.
  Petrini, S.~Raspopovic, G.~Granata, R.~D. Iorio, M.~Controzzi, C.~Cipriani,
  T.~Stieglitz, P.~M. Rossini, and S.~Micera, ``{A closed-loop hand prosthesis
  with simultaneous intraneural tactile and position feedback},'' \emph{Science
  Robotics}, vol.~4, 2019.

\bibitem{Markovic2018}
\BIBentryALTinterwordspacing
M.~Markovic, M.~A. Schweisfurth, L.~F. Engels, D.~Farina, and S.~Dosen,
  ``{Myocontrol is Closed-loop Control: Incidental Feedback is Sufficient for
  Scaling the Prosthesis Force in Routine Grasping},'' \emph{Journal of
  NeuroEngineering and Rehabilitation}, vol.~15, pp. 1--11, 2018.
\BIBentrySTDinterwordspacing

\bibitem{Parry2021AnticipationNeuropathy}
\BIBentryALTinterwordspacing
R.~Parry, F.~R. Sarlegna, N.~Jarrass{\'{e}}, and A.~Roby-Brami, ``{Anticipation
  and Compensation for Somatosensory Deficits in Object Handling: Evidence from
  a Patient with Large Fiber Sensory Neuropathy},'' \emph{Journal of
  Neurophysiology}, vol. 126, pp. 575--590, 2021.
\BIBentrySTDinterwordspacing

\bibitem{Carteron2016TemporaryTasks}
A.~Carteron, K.~McPartlan, C.~Gioeli, E.~Reid, M.~Turturro, B.~Hahn, C.~Benson,
  and W.~Zhang, ``{Temporary Nerve Block at Selected Digits Revealed Hand Motor
  Deficits in Grasping Tasks},'' \emph{Frontiers in Human Neuroscience},
  vol.~0, p. 596, 2016.

\bibitem{Jeannerod1984THELESION}
\BIBentryALTinterwordspacing
M.~Jeannerod, F.~Michel, and C.~Prablanc, ``{The Control of Hand Movements in a
  Case of Hemianaesthesia Following a Parietal Lesion},'' \emph{Brain}, vol.
  107, pp. 899--920, 1984.
\BIBentrySTDinterwordspacing

\bibitem{Hill2009FunctionalGroup}
\BIBentryALTinterwordspacing
W.~Hill, O.~Stavdahl, L.~N. Hermansson, P.~Kyberd, S.~Swanson, and S.~Hubbard,
  ``{Functional outcomes in the WHO-ICF model: Establishment of the upper limb
  prosthetic outcome measures group},'' \emph{Journal of Prosthetics and
  Orthotics}, vol.~21, pp. 115--119, 4 2009.
\BIBentrySTDinterwordspacing

\end{thebibliography}


% \end{IEEEbiography}


% % insert where needed to balance the two columns on the last page with
% %\newpage



%\vfill


\end{document}